\documentclass[twoside]{article}

\usepackage[accepted]{aistats2022}
\usepackage[round]{natbib}

\usepackage{comment}
\usepackage{amsmath}

\usepackage{bm,color,soul}
\usepackage{amsmath,amsfonts,amssymb}
\usepackage{mathtools}
\usepackage[retainorgcmds]{IEEEtrantools}
\usepackage{boldline}
\usepackage{setspace}
\usepackage{bbm}

\usepackage[super]{nth}
\usepackage{dblfloatfix}  
\usepackage{float}
\usepackage[font={footnotesize},labelfont={footnotesize}]{caption}
\usepackage{tikz,pgfplots,adjustbox}
\tikzset{>=latex}
\pgfplotsset{compat=1.3}
\usetikzlibrary{patterns}
\usetikzlibrary{pgfplots.groupplots}

\usepackage{color,soul}

\usepackage{bbold}
\usepgfplotslibrary{fillbetween}

\usepackage{microtype}
\usepackage{graphicx}
\usepackage{subfigure}
\usepackage{booktabs} 

\usepackage{hyperref}
\usepackage{mathtools}
\usepackage{amssymb}
\usepackage{multirow}
\usepackage{multicol}
\usepackage{caption}

\usepackage{color,soul}
\usepackage{xr}
  
\newcommand{\squishlist}{
 \begin{list}{$\bullet$}
  { \setlength{\itemsep}{0pt}
     \setlength{\parsep}{1pt}
     \setlength{\topsep}{1pt}
     \setlength{\partopsep}{0pt}
     \setlength{\leftmargin}{1.5em}
     \setlength{\labelwidth}{1em}
     \setlength{\labelsep}{0.5em} } }
\newcommand{\squishend}{
  \end{list}  }

\newcommand{{\phivmat}}{\boldsymbol{{\phi}}}
\newcommand{{\phiv}}{\boldsymbol{{\phi}}}

\begin{document}

%
\runningtitle{VFDS: Variational Foresight Dynamic Selection in Bayesian Neural Networks}

%
\runningauthor{Randy Ardywibowo, Shahin Boluki,  Zhangyang Wang, Bobak Mortazavi, Shuai Huang, Xiaoning Qian}

\twocolumn[

\aistatstitle{VFDS: Variational Foresight Dynamic Selection in Bayesian Neural Networks for Efficient Human Activity Recognition}

\aistatsauthor{ Randy Ardywibowo$^{1}$ \And Shahin Boluki$^{1}$ \And  Zhangyang Wang$^{2}$ \And Bobak Mortazavi$^{1}$ }
\onehalfspacing
\aistatsauthor{ Shuai Huang$^{3}$ \And Xiaoning Qian$^{1}$ }

\aistatsaddress{ Texas A\&M University$^{1}$ \And University of Texas at Austin$^{2}$ \And University of Washington$^{3}$ } 

]

\begin{abstract}
In many machine learning tasks, input features with varying degrees of predictive capability are acquired at varying costs. In order to optimize the performance-cost trade-off, one would select features to observe \emph{a priori}. However, given the changing context with previous observations, the subset of predictive features to select may change dynamically. Therefore, we face the challenging new problem of \textit{foresight dynamic selection} (FDS): finding a dynamic and light-weight policy to decide which features to observe next, \textbf{before actually observing them}, for overall performance-cost trade-offs. To tackle FDS, this paper proposes a  Bayesian learning framework of \textit{Variational Foresight Dynamic Selection} (\textbf{VFDS}). VFDS learns a policy that selects the next feature subset to observe, by optimizing a variational Bayesian objective that characterizes the trade-off between model performance and feature cost. At its core is an implicit variational distribution on binary gates that are dependent on previous observations, which will select the next subset of features to observe. We apply VFDS on the Human Activity Recognition (HAR) task where the performance-cost trade-off is critical in its practice. Extensive results demonstrate that VFDS selects different features under changing contexts, notably saving sensory costs while maintaining or improving the HAR accuracy. Moreover, the features that VFDS dynamically select are shown to be interpretable and associated with the different activity types. We will release the code.
\end{abstract}

\section{INTRODUCTION}

Acquiring predictive features is critical for building trustworthy machine learning systems, but this may come at a daunting cost. Such a cost can be in the form of energy needed to maintain an ambient sensor \citep{ardywibowo2019adaptive, ardywibowo2018switching, yanginstance}, time needed to complete an experiment \citep{kiefer1959optimum}, or manpower required to monitor a hospital patient \citep{pierskalla1994applications, jiang2019roadmap}. It is important not only to maintain good performance in the specified task, but also a low cost to gather features.

For example, existing Human Activity Recognition (HAR) methods typically use a fixed set of sensors, potentially collecting redundant features to discriminate contexts and/or activity types \citep{shen2013sensor, aziz2016identifying, ertugrul2017determining, cheng2018learning, ardywibowo2017analyzing}. Classic feature selection methods such as the LASSO and its variants can address the performance-cost trade-off by optimizing an objective penalized by a term that helps promote feature sparsity \citep{tibshirani1996regression, friedman2010note, friedman2008sparse, zou2005regularization}. Such feature selection formulations are often static, i.e., a fixed set of features are selected \textit{a priori}. However, different features may offer different predictive power under different contexts. For example, a health worker may not need to monitor a recovering patient as frequently as a patient with declining conditions; or a smartphone sensor may be predictive when the user is walking but not in a car. By dynamically selecting which feature(s) to observe, one can further reduce the inherent cost for prediction and achieve a better trade-off between cost and prediction accuracy.


In addition to cost-efficiency, a dynamic feature selection formulation can also lead to more interpretable and trustworthy predictions. Specifically, the predictions made by the model are only based on the selected features, providing a clear relationship between input features and model predictions. Existing efforts on interpreting models are usually based on some post-analyses of the predictions, including the approaches in (1) visualizing higher-level representations or reconstructions of inputs based on them \citep{li-etal-2016-visualizing, 7299155}, (2) evaluating the sensitivity of predictions to local perturbations of inputs or input gradients \citep{8237336, 10.1145/2939672.2939778}, and (3) extracting parts of inputs as justifications for predictions \citep{lei-etal-2016-rationalizing}. Another related but orthogonal direction is model compression: training sparse neural networks with the goal of memory and computational efficiency \citep{louizos2017learning,10.5555/3327144.3327303, 10.5555/2969239.2969366}. All these works require collecting all features first and provide post-hoc feature or model pruning. 

Recent efforts on dynamic feature selection select which features to observe based on immediate statistics~\citep{gordon2012energy, bloom2013dynamic, ardywibowo2019adaptive, zappi2008activity}, ignoring the information a feature may have on future predictions. Others treat feature selection as a Markov Decision Process (MDP) and use Reinforcement Learning (RL) to solve it \citep{he2012cost, karayev2013dynamic, kolamunna2016afv, spaan2009decision, satsangi2015exploiting, yanginstance}. However, solving RL is not straightforward. Besides being sensitive to hyperparameter settings in general, approximations such as state space discretization and relaxation of the combinatorial objective were used to make the RL problem tractable.

On the other hand, Bayesian inference offers a way to learn a model that formalizes our dynamic feature selection hypothesis. In this direction, \citet{Koop2018BayesianDV} proposed using simple variational distributions to dynamically select predictive models; however, the method is limited to linear, time-varying parameter models. Meanwhile, Bayesian Neural Networks (BNNs) offer a method of training Neural Networks (NNs) while preventing them from overfitting. These methods treat the NN weights as random variables and regularize them with appropriate prior distributions~\citep{mackay1992practical, neal2012bayesian}. To scale these techniques to real-world applications, various types of approximate inference techniques have been developed~\citep{graves2011practical, welling2011bayesian, li2016preconditioned, blundell2015weight, louizos2017multiplicative, shi2017kernel, gal2016dropout, gal2017concrete}. In particular, (semi-)implicit variational inference offers a way to define expressive distributions to better approximate the posterior distribution, enabling more complex models to be inferred efficiently~\citep{yin2018semi, titsias2019unbiased, molchanov2019doubly}. Despite this, the extension of these methods for dynamic feature selection in BNNs has not been explored before. We refer the reader to the supplementary materials for additional related work on static selection, dynamic selection, and variational inference.

To this end, we propose VFDS, a variational dynamic feature selection method for Bayesian Neural Networks that can be easily used with existing deep architecture components and trained from end-to-end, enabling \textit{task-driven dynamic feature selection}. To achieve this, we first define a prior distribution on binary random variables that determines which features to observe next in order to characterize the model performance-cost trade-off. We then design an implicit variational distribution on the binary variables that is conditioned on previous observations, allowing us to dynamically select features at any given time based on previous observations. Through stochastic approximations and differentiable relaxations, we are able to jointly optimize the parameters of this distribution along with the model parameters with respect to the variational objective. To show our method's ability to dynamically select features while maintaining good performance, we evaluate it on four time-series activity recognition datasets: the UCI Human Activity Recognition (HAR) dataset \citep{anguita2013public}, the OPPORTUNITY dataset \citep{roggen2010collecting}, the ExtraSensory dataset \citep{vaizman2017recognizing}, as well as the NTU-RGB-D dataset \citep{Shahroudy_2016_NTURGBD}. 

Several ablation studies and comparisons with other dynamic and static feature selection methods demonstrate the efficacy of our proposed method. In some cases, VFDS only needs to observe 0.28\% features on average while still maintaining competitive human activity monitoring accuracy, compared to 14.38\% used by static feature selection methods. This indicates that some features are indeed redundant for certain contexts. We further show that the dynamically selected features are shown to be interpretable with direct correspondence with different contexts and activity types.




\section{METHODOLOGY}

We define our notation as follows: let $\mathcal{D}$ be a dataset containing $N$ independent and identically distributed~(\emph{iid}) input-output pairs of \emph{multivariate} time series data $\{ (\boldsymbol{x}_1, \boldsymbol{y}_1), \dots, (\boldsymbol{x}_N, \boldsymbol{y}_N) \}$ of length $\{ T_1, \dots, T_N \}$. For each time point $t$, $\boldsymbol{x}_i^t$ is a data point containing $K$ features $\{\boldsymbol{x}_{i,1}^t, \dots, \boldsymbol{x}_{i,K}^t  \}$, and $y_i^t$ is a target output for that time point.

\begin{figure*}[t]
\centering
\subfigure[]{
\includegraphics[width=0.57\linewidth]{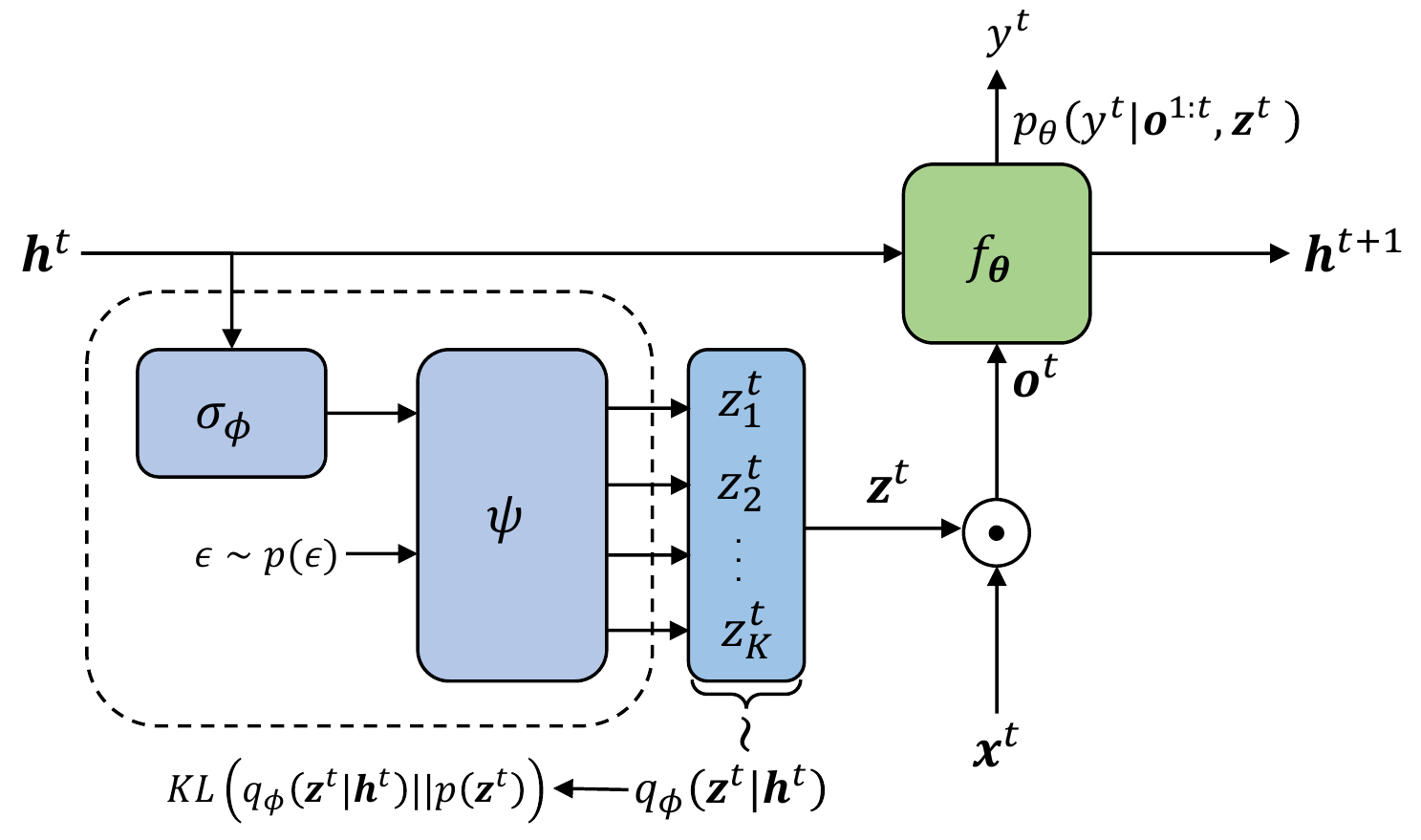}
\label{fig:variationalgating}
}
\subfigure[]{
\includegraphics[width=0.23\linewidth]{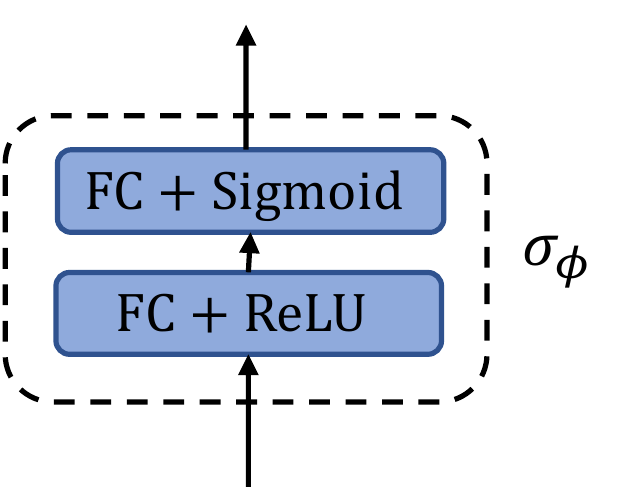}
\label{fig:gatingmodule}
}\vspace{-3mm}
\caption{\textbf{(a)} A variational foresight dynamic feature selection (VFDS) module illustrated for one timestep. $\boldsymbol{h}^t$ is used to determine the variational distribution $q_{\boldsymbol{\phi}}(\boldsymbol{z}^t | \boldsymbol{h}^t)$ from which the feature selection gates $ \boldsymbol{z}^t $ are drawn. $ q_{\boldsymbol{\phi}}(\boldsymbol{z}^t | \boldsymbol{h}^t) $ is defined implicitly through a transformation of a random variable $\boldsymbol{\epsilon}$ drawn from an explicit distribution $p(\boldsymbol{\epsilon})$, by the covariate-dependent function $ \boldsymbol{\psi}_{ \boldsymbol{\phi} }(\boldsymbol{h}^t, \boldsymbol{\epsilon}) $. By choosing $ \boldsymbol{\psi}_{ \boldsymbol{\phi} }(\cdot) $ carefully, we can have $ \mathbb{E}_{p(\boldsymbol{\epsilon})} [ \boldsymbol{\psi}_{ \boldsymbol{\phi} }(\boldsymbol{h}^t, \boldsymbol{\epsilon}) ] = \boldsymbol{\sigma}_{\boldsymbol{\phi}}(\boldsymbol{h}^t) $, where $ \boldsymbol{\sigma}_{\boldsymbol{\phi}}(\cdot) $ can be defined by a neural network. We can obtain a closed form approximation of the KL divergence. The gates are then used to determine which features to observe for the current time-step. \textbf{(b)} The neural network architecture used for $ \boldsymbol{\sigma}_{\boldsymbol{\phi}}(\boldsymbol{h}) $. }
\label{fig:gatingall}
\end{figure*}



We are interested in learning a model that uses the previously observed data to infer the feature set we should observe next, as well as predicting the target output for the next time point. We hypothesize that time-series predictive models for human activity recognition do not require all features be observed at all times. Indeed, many features in multivariate sensor data may be redundant for prediction in a given context. Thus, dynamically choosing which features to observe at any given time would be beneficial as sensors can be dynamically turned on or off depending on specific monitoring needs. Moreover, dynamic feature selection may enable better interpretability on which sensors are required for any given context.

We formalize this hypothesis under the Bayesian Neural Network (BNN) learning paradigm. Deep neural networks offer high predictive capability on complex datasets, allowing us to achieve high performance on the monitoring task. On the other hand, Bayesian statistics offer a way to formalize our hypothesis and learn these neural networks without overfitting. In subsequent sections, we formulate a Bayesian learning problem for foresight dynamic feature selection in terms of variational inference. 


\subsection{Variational Objective of VFDS}
\label{sec:var-obj}

Bayesian learning can be formulated as maximizing a log-marginal likelihood: $ \log {p(\boldsymbol{y} | \boldsymbol{x})} = \log {\prod_{i=1}^{N} {p(\boldsymbol{y}_i | \boldsymbol{x}_i)}} = \log { \int { \prod_{i=1}^{N} { p(\boldsymbol{y}_i | \boldsymbol{x}_i, \boldsymbol{z}) p( \boldsymbol{z} ) d \boldsymbol{z} } } }$, where $ \boldsymbol{z} $ are the intermediary parameters of the model, considered as random variables. This log-marginal is often intractable, and it is common to resort to variational inference by introducing a variational distribution $q(\boldsymbol{z})$ on the parameters of the model. With this, maximizing the log-marginal is often transformed to minimizing the negative Evidence Lower Bound (ELBO)~\citep{hoffman2013stochastic, blei2017variational}. In the case of time-series data with partial observations, the negative ELBO can be written as follows: 
\begin{equation}
\begin{split}
    \mathcal{L}(\mathcal{D}) = - \sum_{i=1}^{N} & { \sum_{t=1}^{T_i} { \mathbb{E}_{\boldsymbol{z} \sim q(\boldsymbol{z}), \boldsymbol{\theta} \sim q(\boldsymbol{\theta}) } [ \log { p (y_i^t | \boldsymbol{o}_i^{1:t}, \boldsymbol{z}, \boldsymbol{\theta}) } ] } } \\ &
    + \mbox{KL}(q(\boldsymbol{z}) || p(\boldsymbol{z})),
\label{eq:elbo}
\end{split}
\end{equation}
where $\boldsymbol{o}_i^{1:t}$ are the observed features up to time $t$. Under this framework, we can form a variational approximation to training dynamic feature selection with deep networks. We do this by introducing stochastic, input-dependent binary variables $z_{i,k}^t$ that determine whether feature $k$ is observed at time $t$. 

To simplify our exposition, we focus on selecting features for one instance $i$ at time-point $t$, and omit these subscripts in our exposition, reintroducing them later for clarity. We present the inference of our binary selection variables. A fully Bayesian treatment of the other neural network model parameters $\boldsymbol{\theta}$ can be considered through standard approximate Bayesian inference techniques for neural networks such as Monte Carlo (MC) dropout and its variants~\citep{gal2016dropout, gal2017concrete, kingma2015variational, pmlr-v108-boluki20a}. For each feature $k$, we define a prior distribution $ p(z_k) $ for each binary variable as 
\begin{subequations}
\begin{eqnarray}
 p(z_k) & = & \mathtt{Bern} ( e^{-\eta c_k } ).
\end{eqnarray}
\end{subequations}
Here, $c_k$ is the energy cost of feature $k$, and $\eta$ is a parameter controlling the shape of the prior. We then define an implicit, covariate-dependent variational distribution that is dependent on a belief state $\boldsymbol{h}$ at time $t$ that summarizes the previous observations. Specifically, the variational distribution $ q (\boldsymbol{z} | \boldsymbol{h}) = q_{\boldsymbol{\phi}} (\boldsymbol{z} | \boldsymbol{h}) $ with parameters $ \boldsymbol{\phi} $ is defined by transforming random variables from an explicit distribution $\boldsymbol{\epsilon} \sim p(\boldsymbol{\epsilon}) $ using a reparameterizable transformation as follows~\citep{kingma2013auto, titsias2019unbiased}:
\begin{equation}
    \boldsymbol{\epsilon} \sim p(\boldsymbol{\epsilon}), \quad \boldsymbol{z} = \boldsymbol{\psi}_{\boldsymbol{\phi}}(\boldsymbol{h}, \boldsymbol{\epsilon}) \quad \equiv \quad \boldsymbol{z} \sim q_{\boldsymbol{\phi}}(\boldsymbol{z} | \boldsymbol{h}).
\end{equation}
Here, $ \boldsymbol{\psi}_{\boldsymbol{\phi}}(\cdot) $ outputs a binary random vector that determines whether feature $k$ is selected, where $ z_{k} = \boldsymbol{\psi}_{\boldsymbol{\phi}}(\boldsymbol{h}, \boldsymbol{\epsilon})_k $. The details of this transformation will be explained in the following sections. 

With this, the first term of $\mathcal{L}(\mathcal{D})$ can be estimated by using a single sample of $\boldsymbol{z}$ for each time point $t$ of instance $i$. On the other hand, the KL term can be computed as
\begin{equation}
    \text{KL}(q(\boldsymbol{z} | \boldsymbol{h}) || p(\boldsymbol{z})) = \sum_{k=1}^{K}{ \text{KL}(q_{\boldsymbol{\phi}}(z_{k} | \boldsymbol{h}) || p(z_{k}) ) },
\end{equation} 
\begin{equation}
\begin{split}
    \text{KL} & (q_{\boldsymbol{\phi}}(z_{k} | \boldsymbol{h}) || p(z_{k}) ) = -H[ q_{\boldsymbol{\phi}}(z_{k} | \boldsymbol{h}) ] + \\ & \eta c_k q_{\boldsymbol{\phi}}(z_{k} = 1 | \boldsymbol{h}) -\log\left( 1-e^{-\eta c_k }\right) q_{\boldsymbol{\phi}}(z_{k} = 0 | \boldsymbol{h}),
\end{split}
\end{equation}
where $ H[ q_{\boldsymbol{\phi}}(z_{k} | \boldsymbol{h}) ] $ is the entropy of $ q_{\boldsymbol{\phi}}(z_{k} | \boldsymbol{h}) $. For sufficiently large $\eta$, $ \log\left( 1-e^{-\eta c_k }\right) \approx 0 $. We achieve this by scaling $\eta$ with $N$, $\eta = N \lambda$. We can then scale the negative ELBO with the number of samples $N$ without changing the optima: 
\begin{equation}
\begin{split}
    \text{KL} & (q_{\boldsymbol{\phi}}(z_{k} | \boldsymbol{h}) || p(z_{k}) ) \approx \\ & - \frac{1}{N} H[ q_{\boldsymbol{\phi}}(z_{k} | \boldsymbol{h}) ] + \frac{N\lambda}{N} c_k q_{\boldsymbol{\phi}}(z_{k} = 1 | \boldsymbol{h}).
\end{split}
\end{equation}
For large $N$, the entropy term vanishes, leaving us with 
\begin{equation}
    \text{KL}(q_{\boldsymbol{\phi}}(z_{k} | \boldsymbol{h}) || p(z_{k}) ) \approx \lambda c_k  q_{\boldsymbol{\phi}}(z_{k} = 1 | \boldsymbol{h}).    
\end{equation}
Note that $ q_{\boldsymbol{\phi}}(z_{k} = 1 | \boldsymbol{h}) =  \mathbb{E}_{p(\boldsymbol{\epsilon})} [ \boldsymbol{\psi}_{\boldsymbol{\phi}}(\boldsymbol{h}, \boldsymbol{\epsilon})_k ] $. By defining $ p(\boldsymbol{\epsilon}) $ as the logistic distribution with probability density $f(\boldsymbol{\epsilon})$, and $ \boldsymbol{\psi}_{ \boldsymbol{\phi} }(\boldsymbol{h}, \boldsymbol{\epsilon}) $ through a deterministic function $ \boldsymbol{\sigma}_{\boldsymbol{\phi}}(\boldsymbol{h}) \in ( 0, 1 ) $ as follows:
\begin{equation}
    \boldsymbol{\epsilon} \sim p(\boldsymbol{\epsilon}), \quad f(\boldsymbol{\epsilon}) = \frac{e^{-\boldsymbol{\epsilon}}}{(1+e^{-\boldsymbol{\epsilon}})^2},
\end{equation}
\begin{equation}
    \boldsymbol{\psi}_{ \boldsymbol{\phi} }(\boldsymbol{h}, \boldsymbol{\epsilon}) = \mathbbm{1} \bigg[ \log \bigg( \frac{\boldsymbol{\sigma}_{\boldsymbol{\phi}}(\boldsymbol{h})}{1-\boldsymbol{\sigma}_{\boldsymbol{\phi}}(\boldsymbol{h})} \bigg) + \boldsymbol{\epsilon} > 0 \bigg],
\end{equation}
we have that $ \mathbb{E}_{p(\boldsymbol{\epsilon})}[ \boldsymbol{\psi}_{ \boldsymbol{\phi} }(\boldsymbol{h}, \boldsymbol{\epsilon}) ] = \boldsymbol{\sigma}_{\boldsymbol{\phi}}(\boldsymbol{h}) $. In practice, $ \boldsymbol{\epsilon} \sim p(\boldsymbol{\epsilon}) $ can be sampled as $ \boldsymbol{\epsilon} = \log{\boldsymbol{u}} - \log(1-\boldsymbol{u}) $, where $\boldsymbol{u} \sim \mathtt{Unif}(0, 1)$. On the other hand, we specify $ \boldsymbol{\sigma}_{\boldsymbol{\phi}}(\boldsymbol{h}) $ as a neural network, whose architecture we will describe in later sections. Now, our approximation to the KL term can be written as 
\begin{equation}
    \text{KL}(q_{\boldsymbol{\phi}}(z_{k} | \boldsymbol{h}) || p(z_{k}) ) \approx \lambda c_k \boldsymbol{\sigma}_{\boldsymbol{\phi}}(\boldsymbol{h})_k.
\end{equation}
By using this approximation in the negative ELBO $\mathcal{L}(\mathcal{D})$, rewriting the ELBO in terms of an expectation, and reintroducing subscripts, we arrive at the following objective: 
\begin{equation}
\begin{split}
    \mathcal{L} (\mathcal{D}) & = \mathbb{E}_{(\boldsymbol{x}, \boldsymbol{y}, i) \sim \mathcal{D}} \bigg[ \\ &  - \sum_{t=1}^{T_i} { \mathbb{E}_{\boldsymbol{z}^t_i \sim q_{ \boldsymbol{\phi} }(\boldsymbol{z}^t_i | \boldsymbol{h}^t_i), \boldsymbol{\theta} \sim q_{ \boldsymbol{\phi} }(\boldsymbol{\theta})} \big[
    \log {p (y^t | \boldsymbol{x}^t, \boldsymbol{z}^t_i, \boldsymbol{\theta}) } \big] } \\ & +
    \lambda \sum_{t=1}^{T_i} { \sum_{k=1}^{K}{ c_k \boldsymbol{\sigma}_{\boldsymbol{\phi}}(\boldsymbol{h}^t_i)_k } } \bigg].
\end{split}
\end{equation} 
We see here that there are two terms in our objective function, the first term is a likelihood term that determines how accurately the model recovers the target distribution, and the second term penalizes the model for dynamically choosing to observe too many features on average at each time-point, weighted by their energy cost. Intuitively, $ \boldsymbol{\sigma}_{\boldsymbol{\phi}}(\boldsymbol{h}) $ can be thought of as a gating module that selects features to observe based on the previous observations. In the following sections, we describe the architecture of this gating mechanism in greater details, as well as practical considerations when attempting to optimize with respect to this objective and apply this method in practice.

\subsection{Foresight Dynamic Selection Module}

We now describe our dynamic feature selection module, which can be seen in Figure~\ref{fig:gatingall}. Here, we adopt a Recurrent Neural Network (RNN) structure, using $f_{\boldsymbol{\theta}}(\cdot)$ with parameters $\boldsymbol{\theta}$ to compute the belief state $\boldsymbol{h}^t$ for a given time $t$~\citep{graves2013speech}. We then use the hidden state $\boldsymbol{h}$ to implicitly define the distribution $q_{\boldsymbol{\phi}}(\boldsymbol{z}^t | \boldsymbol{h}^t)$ from which we sample the feature selection gates $\boldsymbol{z}^t$. This is done by first feeding $\boldsymbol{h}^t$ through a gating module $\sigma_{\boldsymbol{\phi}}(\boldsymbol{h}^t)$ with variational parameters $\boldsymbol{\phi}$. This module is defined by a neural network consisting of two fully connected layers with ReLU and sigmoid activation functions,  respectively. This can be seen in Figure~\ref{fig:gatingmodule}. We then use the output of this module to transform the random variable $\epsilon$ into $\boldsymbol{z}^t$, thereby sampling $\boldsymbol{z}^t$ from $ q_{\boldsymbol{\phi}}(\boldsymbol{z}^t | \boldsymbol{h}^t) $.

With this, our optimization problem aims at minimizing $\mathcal{L}(\mathcal{D})$ by optimizing the parameters $\boldsymbol{\theta}$ and variational parameters $\boldsymbol{\phi}$. We intend to solve this problem through gradient-based methods. However, the discrete random variables $ \boldsymbol{z}^t $'s are not directly amenable to stochastic reparameterization techniques. In the following, we describe a differentiable relaxation that we adopt to allow the training of our method end-to-end, enabling easy integration into many existing deep architectures.


\subsection{Differentiable Relaxation}

The final hurdle in solving the above problem using gradient descent is that the discrete random variables $ \boldsymbol{z}^t $'s are not directly amenable to stochastic reparameterization techniques. An effective and simple to implement formulation that we adopt is the Gumbel-Softmax reparameterization \citep{jang2016categorical, maddison2016concrete, ardywibowo2020nads}. It relaxes a discrete valued random variable $\boldsymbol{z}$ to a continuous random variable $\tilde{\boldsymbol{z}}$. Specifically, the discrete valued random variables $\boldsymbol{z}$ can instead be relaxed into continuous random variables $\tilde{\boldsymbol{z}}$ through the transformation $ \tilde{\boldsymbol{\psi}}_{ \boldsymbol{\phi} }(\boldsymbol{x}, \boldsymbol{\epsilon}) $ as follows: 
\begin{equation}
\begin{split}
    \tilde{\boldsymbol{\psi}}_{ \boldsymbol{\phi} }(\boldsymbol{x}, \boldsymbol{\epsilon}) = \textsc{Sigmoid} \bigg( \bigg( \log \bigg( \frac{\boldsymbol{\sigma}_{\boldsymbol{\phi}}(\boldsymbol{x})}{1-\boldsymbol{\sigma}_{\boldsymbol{\phi}}(\boldsymbol{x})} \bigg) + \boldsymbol{\epsilon} \bigg) \bigg/ \tau \bigg),
\end{split}
\label{eq:gumbel}
\end{equation}
where $ \boldsymbol{\epsilon} $ is a sample from a logistic distribution defined in Section~\ref{sec:var-obj}. Meanwhile, $\tau$ is a temperature parameter. For low values of $\tau$, $ \boldsymbol{s}$ approaches a sample of a binary random variable, recovering the original discrete problem, while for high values, $ \boldsymbol{s}$ will equal $\frac{1}{2}$.

With this, we are able to compute gradient estimates of $\tilde{\boldsymbol{z}}$ and approximate the gradient of $\boldsymbol{z}$ as $\nabla_{\boldsymbol{\theta}, \boldsymbol{\phi}} \boldsymbol{z} \approx \nabla_{\boldsymbol{\theta}, \boldsymbol{\phi}} \tilde{\boldsymbol{z}} $. This enables us to backpropagate through the discrete random variables and train the selection parameters along with the model parameters jointly using stochastic gradient descent. At test time, we remove the stochasticity and set the gates as $ \boldsymbol{z} = \mathbbm{1} \big[ \log \big( \frac{\boldsymbol{\sigma}_{\boldsymbol{\phi}}(\boldsymbol{x})}{1-\boldsymbol{\sigma}_{\boldsymbol{\phi}}(\boldsymbol{x})} \big) > 0 \big] $, or equivalently, set $ \boldsymbol{z} = \mathbbm{1} [ \boldsymbol{\sigma}_{\boldsymbol{\phi}}(\boldsymbol{x}) > \frac{1}{2} ] $.

We can see that such a module can be easily integrated into many existing deep architectures and trained from end-to-end, enabling \textit{task-driven feature selection}. We demonstrate this ability by applying it to a variety of recurrent architectures such as a Gated Recurrent Unit (GRU)~\citep{cho2014learning} and an Independent RNN~\citep{li2018independently}. 


\section{EXPERIMENTS}

 

We evaluate VFDS on four different datasets: the UCI Human Activity Recognition (HAR) using Smartphones Dataset \citep{anguita2013public}, the OPPORTUNITY Dataset \citep{roggen2010collecting}, the ExtraSensory dataset \citep{vaizman2017recognizing}, and the NTU-RGB-D dataset \citep{Shahroudy_2016_NTURGBD}. Although there are many other human activity recognition benchmark datasets \citep{chen2020deep}, we choose the above datasets to better convey our message of achieving feature usage efficiency and interpretability using our dynamic feature selection framework with the following reasons. First, the UCI HAR dataset is a clean dataset with no missing values, allowing us to benchmark different methods without any discrepancies in data preprocessing confounding our evaluations. Second, the OPPORTUNITY dataset contains activity labels that correspond to specific sensors. An optimal dynamic feature selector should primarily choose these sensors under specific contexts with clear physical meaning. The ExtraSensory dataset studies a multilabel classification problem, where two or more labels can be active at any given time. Finally, the NTU-RGB-D dataset is a large-scale activity recognition dataset with over 60 classes of activities using data from 25 skeleton joints, allowing us to benchmark model performance in a complex setting. For all datasets, we randomly split data both chronologically and by different subjects.

\begin{table}[b]
\vspace{-3mm}
\caption{Comparison of various optimization techniques for our model on the UCI HAR dataset. {\footnotesize *Accuracy and average number of features selected are in (\%).}}
\label{table:optim}
\begin{center}
\begin{tabular}{lcc}
\toprule
Method & Accuracy & Feat. Selected \\ 
\hline
$\ell_1$ Regularization & 90.43 & 19.48 \\
Straight Through & 89.38 & 0.31 \\
ARM & 95.73 & 11.67 \\
ST-ARM & 92.79 & 1.92 \\
Gumbel-Softmax & \textbf{97.18} & \textbf{0.28} \\
\bottomrule
\end{tabular}
\end{center}
\end{table}


\begin{table*}[t]
\caption{Comparison of various models for dynamic feature selection on three activity recognition datasets. {\footnotesize *Accuracy metrics and average number of features selected are all in (\%).}}
\centering
\makebox[\linewidth]{
\begin{tabular}{lcc|cc|ccc}
\toprule
\multirow{2}{*}{Method} & \multicolumn{2}{c}{UCI HAR} & \multicolumn{2}{c}{OPPORTUNITY} & \multicolumn{3}{c}{ExtraSensory} \\
\cline{2-8}
& Accuracy & Features & Accuracy & Features & Accuracy & F1 & Features \\
\hline
No Selection (GRU) & 96.67 & 100 & \textcolor{blue}{84.16} & 100 & \textbf{\textcolor{red}{91.14}} & \textcolor{blue}{53.53} & 100 \\
\hline
Static & 95.49 & 14.35 & 81.63 & 49.57 & 91.13 & 53.18 & 42.32 \\
Random & 52.46 & 25.00 & 34.11 & 50.00 & 39.66 & 23.53 & 40.00 \\
MDP & 62.21 & 24.58 & 44.45 & 34.68 & 48.20 & 28.11 & 31.98 \\
IDSS & 87.88 & \textcolor{blue}{10.39} & 72.95 & \textcolor{blue}{28.50} & 70.59 & 33.24 & \textcolor{blue}{22.07} \\
Attention & \textbf{\textcolor{red}{98.38}} & 49.94 & 83.42 & 54.20 & 90.37 & 53.29 & 54.73 \\
\hline
VFDS & \textcolor{blue}{97.18} & \textbf{\textcolor{red}{0.28}} & \textbf{\textcolor{red}{84.26}} & \textbf{\textcolor{red}{15.88}} & \textbf{\textcolor{red}{91.14}} & \textbf{\textcolor{red}{55.06}} & \textbf{\textcolor{red}{11.25}} \\
\bottomrule
\end{tabular}
}
\label{table:all}
\end{table*}

We investigate several aspects of our model performance on these benchmarks. To show the effect in prediction accuracy when our selection module is considered, we compare its performance to a standard GRU architecture~\citep{cho2014learning}. To show the effect of considering dynamic feature selection, we compare a static feature selection formulation using the technique by \citet{louizos2017learning}. To benchmark the performance of our differentiable relaxation-based optimization strategy, we implement the Straight-Through estimator~\citep{hinton2012neural}, $\ell_1$ relaxed regularization, and Augment-REINFORCE-Merge~(ARM) gradient estimates~\citep{yin2018arm} as alternative methods to optimize our formulation. The fully sequential application of ARM was not addressed in the original paper, and will be prohibitively expensive to compute exactly. Hence, we combine ARM and Straight-Through (ST) estimator \citep{hinton2012neural} as another approach to optimize our formulation. More specifically, we calculate the gradients with respect to the Bernoulli variables with ARM, and use the ST estimator to backpropagate the gradients through the Bernoulli variables to previous layers' parameters. We further compare with an attention-based feature selection, selecting features based on the largest attention weights. Because attention yields feature attention weights instead of feature subsets, we select features by using a hard threshold $\alpha$ of the attention weights and scaling the selected features by $1-\alpha$ for different values of $\alpha$. Indeed, without this modification, we observe that an attention-based feature selection would select 100\% of the features at all times. As additional benchmarks, we compare against a random selection baseline, a Markov Decision Process (MDP), and IDSS, a Reinforcement Learning (RL) based method by~\citet{yanginstance}.


We also have tested different values for the temperature hyperparameter $\tau$, where we observe that the settings with $\tau$ below 1 generally yield the best results with no noticeable performance difference. This experiment, discussions on the ExtraSensory dataset, and additional experiments on the stability of our dynamic selection formulation can be found in the supplementary materials.




\textbf{UCI HAR Dataset}: We test our proposed method on performing simultaneous prediction and dynamic feature selection on the UCI HAR dataset \citep{anguita2013public}. This dataset consists of 561 smartphone sensor measurements including various gyroscope and accelerometer readings, with the task of inferring the activity that the user performs at any given time. There are six possible activities that a subject can perform: walking, walking upstairs, walking downstairs, sitting, standing, and laying. Additional experiment details can be found in the supplementary materials.

\begin{figure*}[t]
\centering
\subfigure[]{
\includegraphics[width=0.33\textwidth]{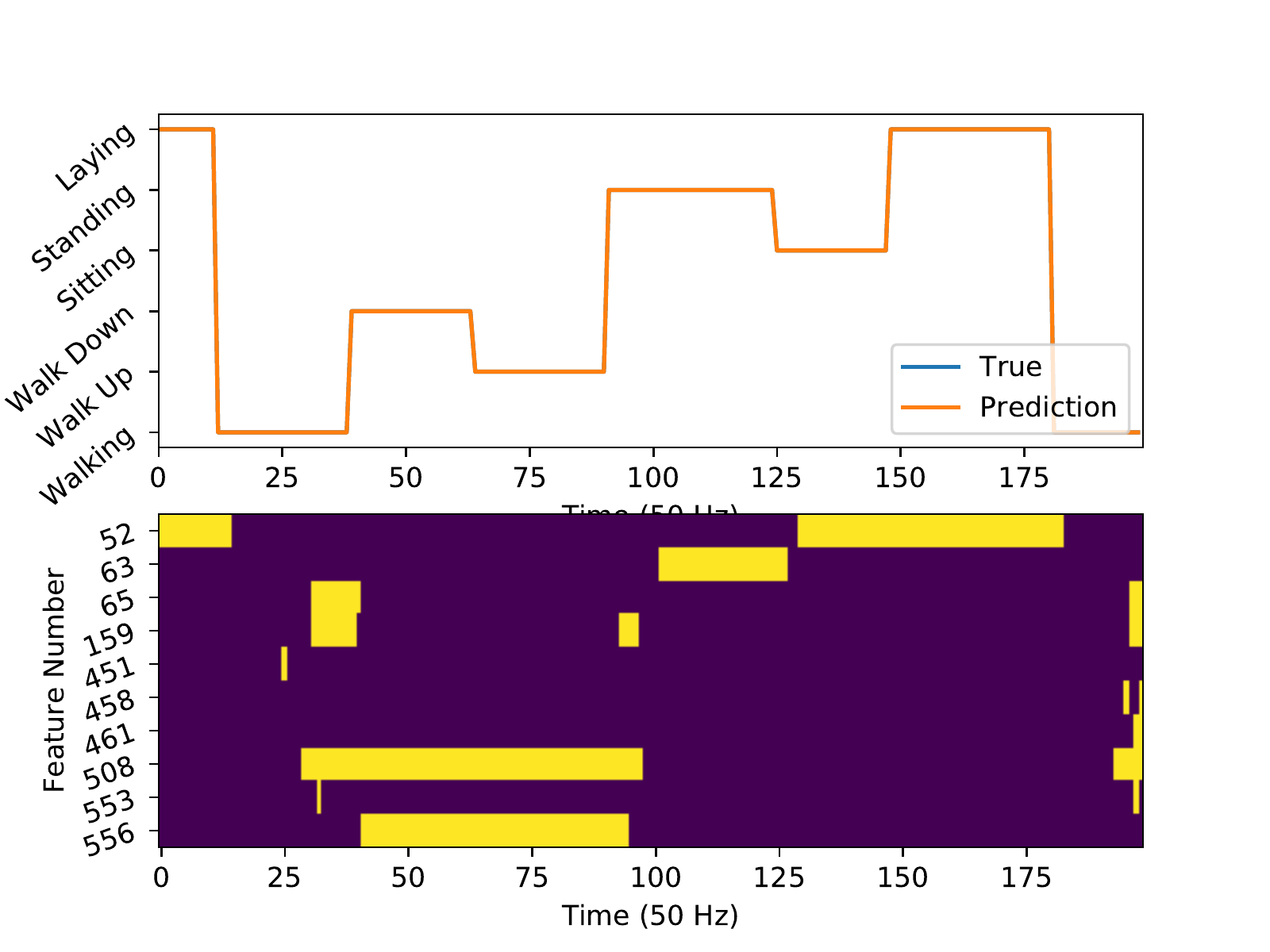}
\label{fig:traj}
}
\subfigure[]{
\includegraphics[width=0.23\textwidth]{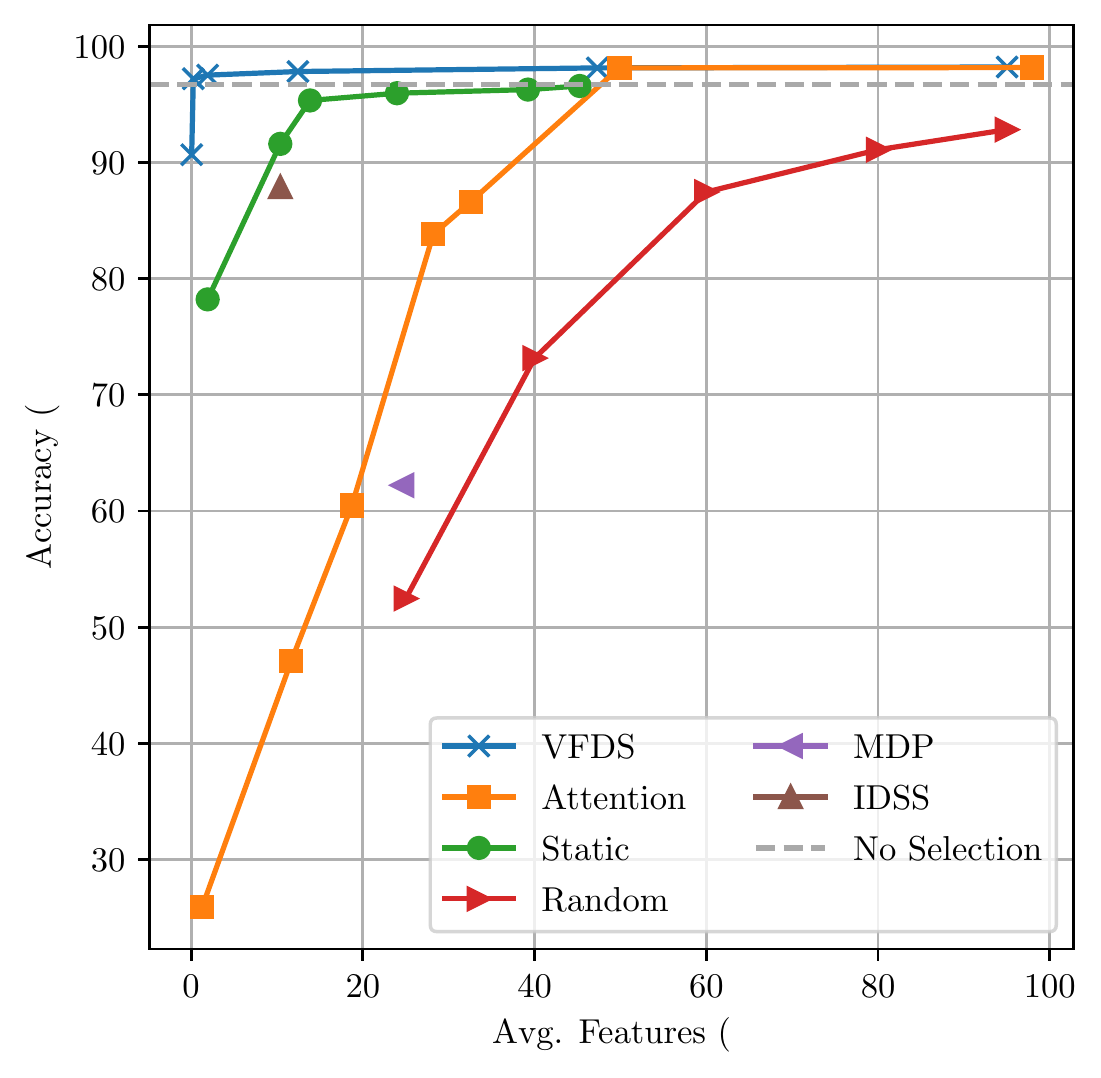}
\label{fig:tradeoff}
}
\subfigure[]{
\includegraphics[width=0.38\textwidth]{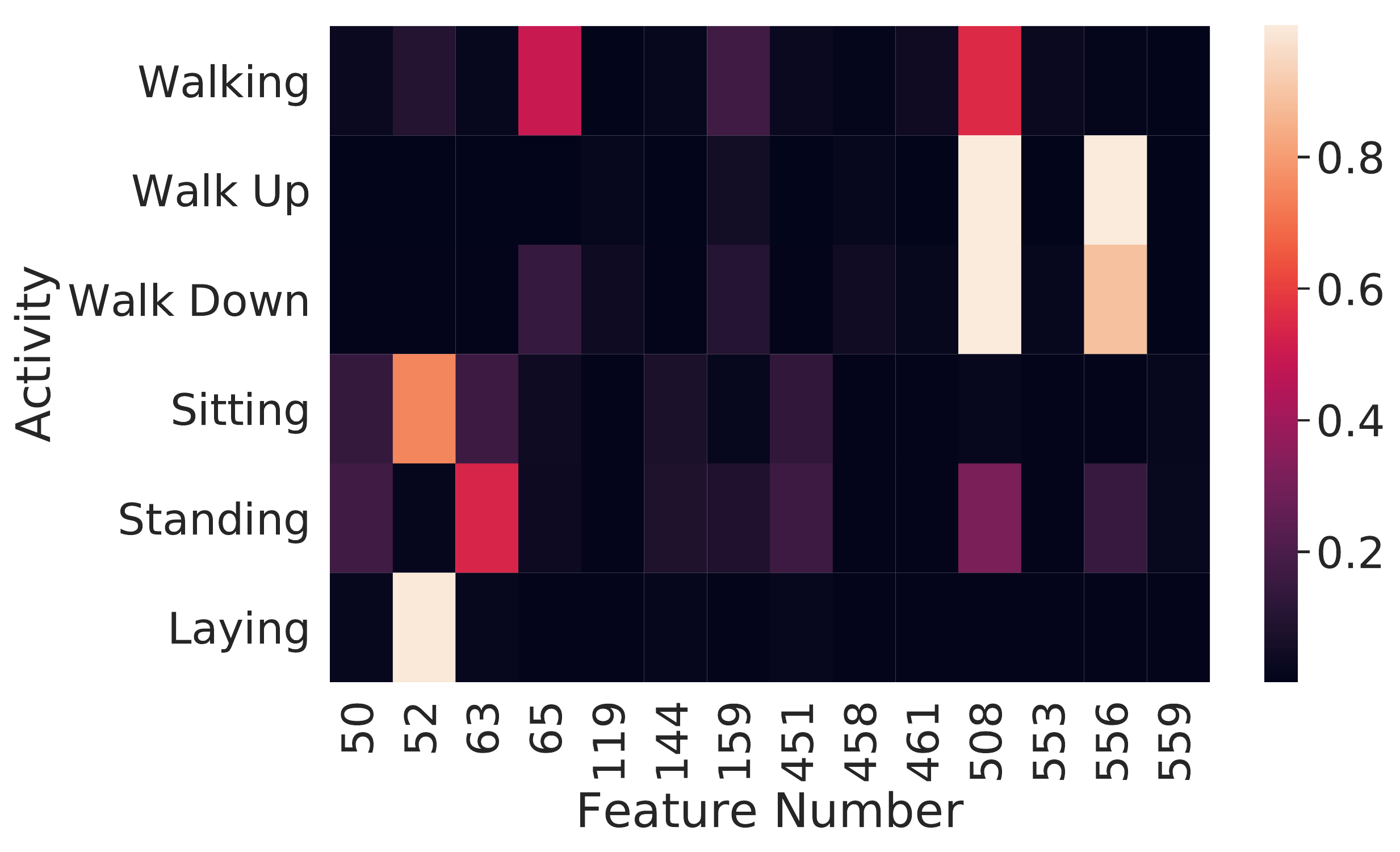}
\label{fig:heatmap-har}
}
\caption{UCI HAR Dataset results: \textbf{(a)} Prediction and features selected of the proposed model $\lambda=1$. \textbf{(b)} Feature selection vs. accuracy trade-off curve comparison. \textbf{(c)} Heatmap of sensor feature activations under each activity of the UCI HAR dataset. Only active features are shown out of the 561 features in total.}
\vspace{-3mm}
\end{figure*}

We first compare various optimization methods using stochastic gradients by differential relaxation via Gumbel-Softmax (GS) reparametrization, Straight-Through (ST), ARM, ST-ARM gradients, and an $\ell_1$ regularized formulation to solve dynamic feature selection. As shown by the results provided in Table~\ref{table:optim}, Gumbel-Softmax achieves the best prediction accuracy with the least number of features. Utilizing either ST, ARM, or ST-ARM for gradient estimation cannot provide a better balance between accuracy and efficiency compared with the Gumbel-Softmax relaxation-based optimization. Indeed, the performance of the ST estimator is expected, as there is a mismatch between the forward propagated activations and the backward propagated gradients in the estimator. Meanwhile, we attribute the worse performance of the ARM and ST-ARM optimizer to its use in a sequential fashion, which was not originally considered. The lower performance of the $\ell_1$ regularized formulation is expected as $\ell_1$ regularization is an approximation to the optimal feature subset selection problem.

Benchmarking results of different models are given in Table~\ref{table:all}. As shown, our dynamic feature selection model is able to achieve a competitive accuracy using only 0.28\% of the features, or on average about 1.57 sensors at any given time. We also observe that both the attention and our dynamic formulation are able to improve upon the accuracy of the standard GRU, suggesting that feature selection can also regularize the model to improve accuracy. Although the attention-based model yields the best accuracy, about 50\% of the features are used on average compared to 0.28\% for our method.

\begin{figure*}[b]
\vspace{-3mm}
\centering
\subfigure[]{
\includegraphics[width=0.34\textwidth]{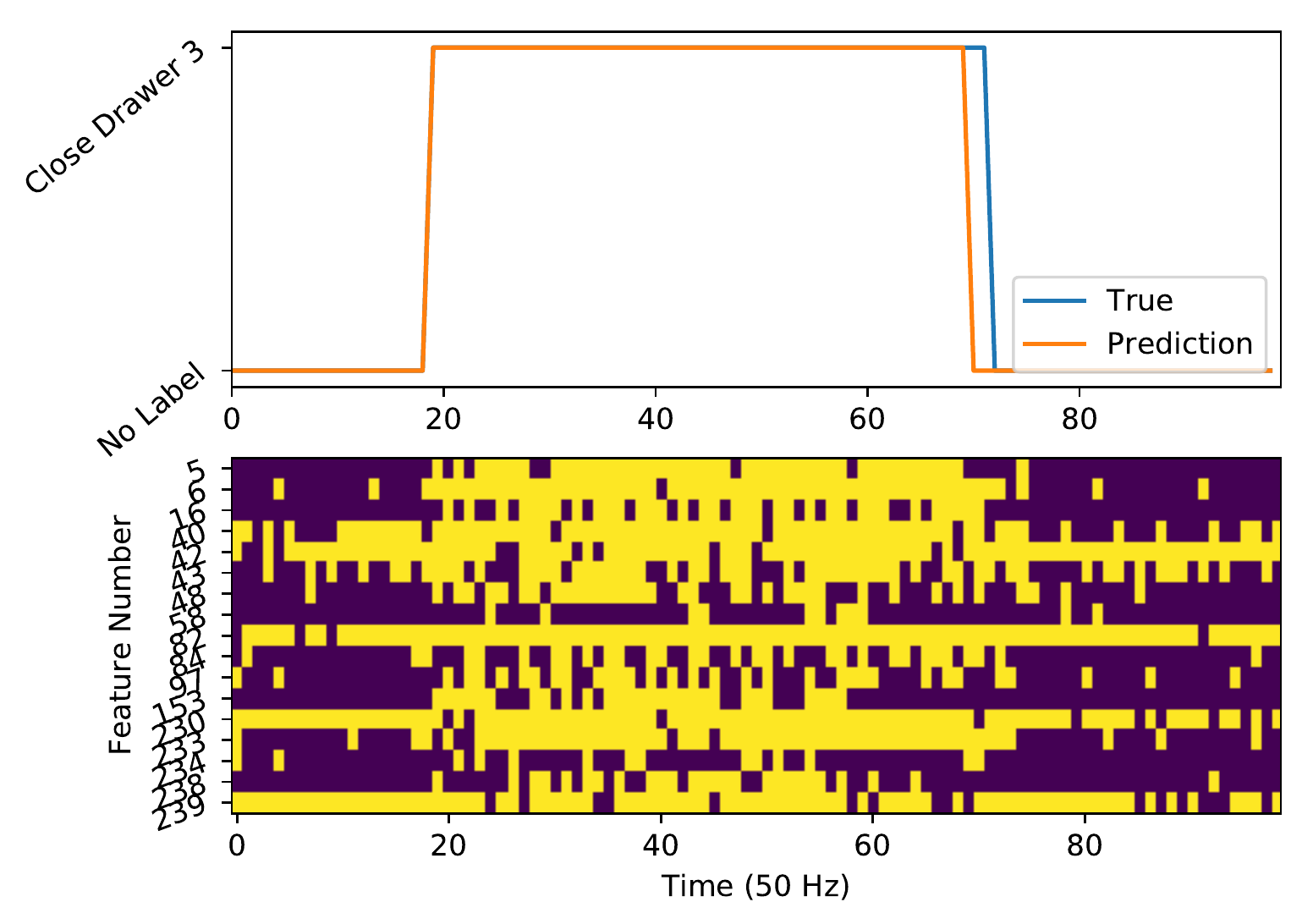}
\label{fig:traj-oppor1}
}
\subfigure[]{
\includegraphics[width=0.34\textwidth]{opportunity_trajectory3.pdf}
\label{fig:traj-oppor2}
}
\subfigure[]{
\includegraphics[width=0.25\textwidth]{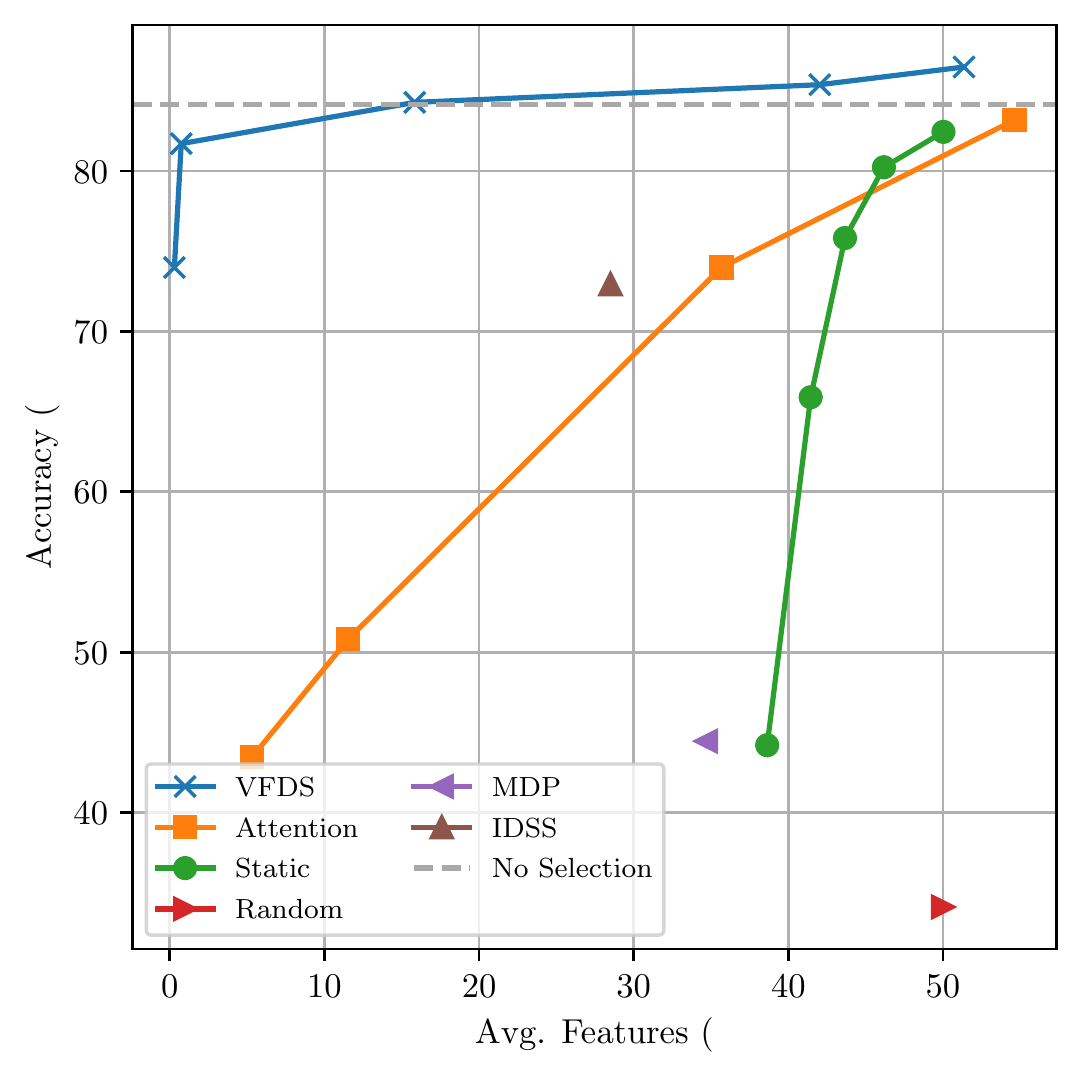}
\label{fig:tradeoff-oppor}
}
\caption{OPPORTUNITY Dataset results: \textbf{(a)} Prediction and features selected of the proposed model $\lambda=1$. \textbf{(b)} Prediction and features selected of the proposed model on a set of activity transitions. \textbf{(c)} Feature selection vs. Error trade-off curve comparison.}
\end{figure*}

We study the effect of the regularization weight $\lambda$ by varying it from $\lambda \in \{ 1, 0.1, 0.01, 0.005, 0.001 \}$. We compare this with the attention model by varying the threshold $\alpha$ used to select features from $\alpha \in \{ 0.5, 0.9, 0.95, 0.99, 0.995, 0.999 \}$, as well as the static selection model by varying its $\lambda$ from $\lambda \in \{ 1, 0.1, \dots 0.01, 0.005, 0.001 \}$. A trade-off curve between the number of selected features and the performance for the three models can be seen in Figure~\ref{fig:tradeoff}. As shown in the figure, the accuracy of the attention model suffers increasingly with smaller feature subsets, as attention is not a formulation specifically tailored to find sparse solutions. On the other hand, the accuracy of our dynamic formulation is unaffected by the number of features, suggesting that selecting around 0.3\% of the features on average may be optimal for the given problem. It further confirms that our dynamic formulation selects the most informative features given the context. Moreover, as we show in the supplementary materials, some features are not selected at all by our dynamic feature selector. The performance of the static selection model is consistent for feature subsets of size 10\% or greater. However, it suffers a drop in accuracy for extremely small feature subsets. This shows that for static selection, selecting too many features may result in collecting redundant ones for certain contexts, while selecting a feature set that is too small would be insufficient for maintaining accuracy.

\begin{figure*}[t]
\centering
\includegraphics[width=0.85\textwidth]{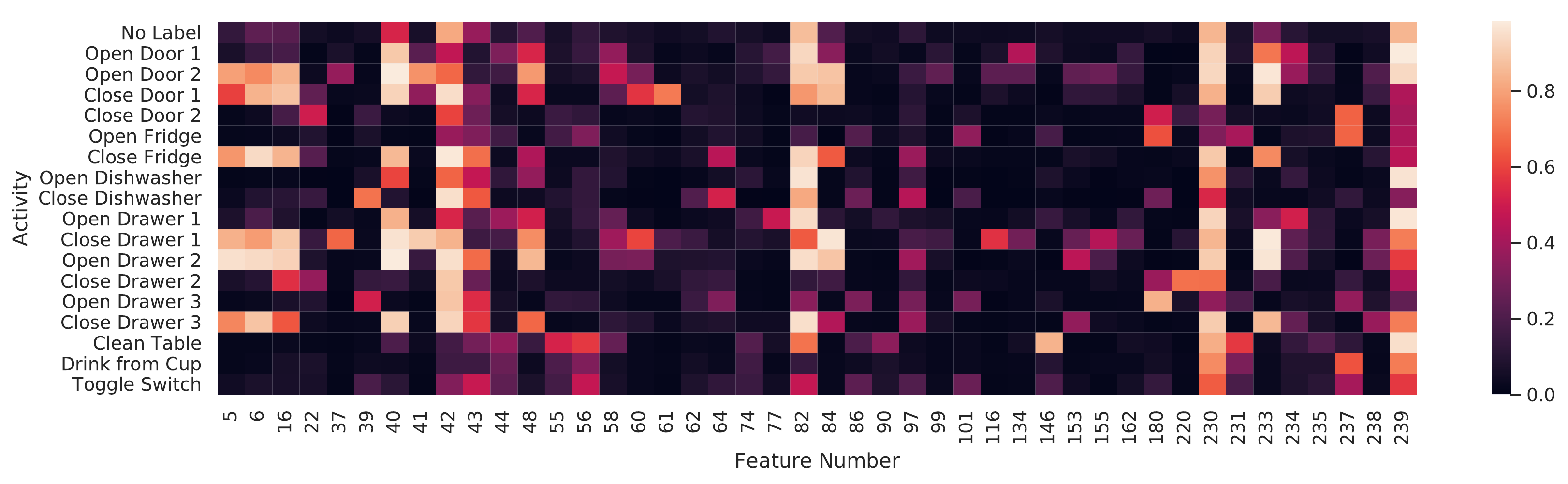}
\caption{Heatmap of sensor feature activations under each activity of the OPPORTUNITY dataset. \footnotesize *Only active features are shown out of the 242 features in total.}
\label{fig:heatmap-oppor}
\vspace{-3mm}
\end{figure*}

An example of dynamically selected features can be seen in Figure~\ref{fig:traj}. We plot the prediction of our model compared to the true label and illustrate the features that are used for prediction. We also plot a heatmap for the features selected under each activity in Figure~\ref{fig:heatmap-har}. Although these features alone may not be exclusively attributed as the only features necessary for prediction under specific activities, such a visualization is useful to retrospectively observe the features selected by our model at each time-point. Note that mainly 5 out of the 561 features are used for prediction at any given time. Observing the selected features, we see that for the static activities such as sitting, standing, and laying, only sensor feature 52 and 63, features relating to the gravity accelerometer, are necessary for prediction. On the other hand, the active states such as walking, walking up, and walking down requires 3 sensor features: sensor 65, 508, and 556, which are related to both the gravity accelerometer and the body accelerometer. This is intuitively appealing as, under the static contexts, the body accelerometer measurements would be relatively constant, and unnecessary for prediction. On the other hand, for the active contexts, the body accelerometer measurements are necessary to reason about how the subject is moving and accurately discriminate between the different active states. Meanwhile, we found that measurements relating to the gyroscope were unnecessary for prediction. 



\textbf{OPPORTUNITY Dataset}: We further test our proposed method on the UCI OPPORTUNITY Dataset \citep{roggen2010collecting}. This dataset consists of multiple different label types for human activity, ranging from locomotion, hand gestures, to object interactions. The dataset consists of 242 measurements from accelerometers and Inertial Measurement Units (IMUs) attached to the user, as well as accelerometers attached to different objects with which the user can interact. Additional experiment details can be found in the supplementary materials.

We use the mid-level gesture activities as the target for our models to predict, which contain gestures related to specific objects, such as opening a door and drinking from a cup. A comparison of the accuracy and the percentage of selected features by different models is given in Table~\ref{table:all}, while example predictions and a trade-off curve are constructed and shown in Figures \ref{fig:traj-oppor1}, \ref{fig:traj-oppor2}, and \ref{fig:tradeoff-oppor}, with a similar trend as the results on the UCI HAR dataset.

A heatmap for the selected features under each activity is shown in Figure~\ref{fig:heatmap-oppor}. Here, the active sensor features across all activities are features 40 and 42, readings of the IMU attached to the subject's back, feature 82, readings from the IMU attached to the Left Upper Arm (LUA), and features 230 and 239, location tags that estimate the subject's position. We posit that these general sensor features are selected to track the subject's overall position and movements, as they are also predominantly selected in cases with no labels. Meanwhile, sensors 5, 6, and 16, readings from the accelerometer attached to the hip, LUA, and back, are specific to activities involving opening/closing doors or drawers.

\textbf{NTU-RGB-D Dataset}: We further test our proposed method on the NTU-RGB-D dataset~\citep{Shahroudy_2016_NTURGBD}. This dataset consists of 60 different activities performed by either a single individual or two individuals. The measurements of this dataset are in the form of skeleton data consisting of 25 different 3D coordinates of the corresponding joints of the participating individuals. Additional experiment details can be found in the supplementary materials.

We compare our method with three different baselines shown in Table~\ref{table:optim-ntu}: the baseline RNN architecture, soft attention, and thresholded attention baseline. We see that our method maintains a competitive accuracy compared to the baseline using less than 50\% of the features. On the other hand, because the thresholded attention formulation is not specifically optimized for feature sparsity, we see that it performs significantly worse compared to the other methods. Meanwhile, the soft-attention slightly improves upon the accuracy of the base architecture. However, as also indicated by our other experiments, soft-attention is not a dynamic feature selection method, and tends to select 100\% of the features at all times.

The results on these four datasets indicate that our dynamic monitoring framework provides the best trade-off between feature efficiency and accuracy, while the features that it dynamically selects are also interpretable and associated with the actual activity types.

\begin{table}[t]
\caption{Comparison of various methods for activity recognition on the NTU-RGB-D dataset. *Accuracy and average number of features selected are in (\%).}
\label{table:optim-ntu}
\begin{center}
\begin{tabular}{lcc}
\toprule
Method & Accuracy (\%) & Features (\%) \\ 
\hline
No Selection & 83.02 & 100 \\
Soft Attention & 83.28 & 100 \\
Thresh. Attention & 40.07 & 52.31 \\
\hline
VFDS & \textbf{83.31} & \textbf{49.65} \\
\bottomrule
\end{tabular}
\end{center}
\vspace{-3mm}
\end{table}




\section{CONCLUSIONS}
We have introduced a novel method, VFDS, for performing foresight dynamic feature selection through variational inference. We accomplish this by defining a variational objective with a prior that captures the performance-cost trade-off of observing a given feature at a given time-point. We then designed an implicit, covariate-dependent variational distribution, and use a differentiable relaxation, making the optimization amenable to stochastic gradient-based optimization. As our method is easily applicable to existing neural network architectures, we are able to apply our method on Recurrent Neural Networks for human activity recognition. We benchmark our model on four different activity recognition datasets and have compared it with various dynamic and static feature selection benchmarks. Our results show that our model maintains a desirable prediction performance using a significantly small fraction of the sensors or features. The features that our model selected were shown to be interpretable and associated with the activity types.

\section*{Acknowledgments} 

This project is supported in part by the Defense Advanced Research Projects Agency (DARPA) under grant FA8750-18-2-0027. X. Qian has been supported in part by the National Science Foundation (NSF) Awards 1553281, 1812641, 1835690, 1934904, and 2119103. 

\bibliography{references}
\bibliographystyle{aaai}




\end{document}


%

%

\onecolumn
\aistatstitle{\textbf{Supplementary Material}: \\
VFDS: Variational Foresight Dynamic Selection in Bayesian Neural Networks for Efficient Human Activity Recognition}

\section{Related Work}


\subsection{Sensor Selection in Human Activity Recognition}

Existing HAR systems typically use a fixed set of sensors, potentially collecting redundant features for easily discriminated contexts. Methods that attempt to find a fixed or static feature set often rank feature sets using metrics such as Information Gain \citep{shen2013sensor}, or relevancy ranking through a filtering strategy \citep{aziz2016identifying, ertugrul2017determining, cheng2018learning}. However, static feature selection can potentially result in collecting redundant information for highly distinguishable contexts.

\subsection{Dynamic Feature Selection}

Work on dynamic feature selection can be divided into Reinforcement Learning (RL) based and non-RL approaches. Non-RL based approaches vary from assigning certain features to certain activities \citep{gordon2012energy}, pre-defining feature subsets for prediction \citep{bloom2013dynamic, strubell2015learning}, optimizing the trade-off between prediction entropy and the number of selected features  \citep{ardywibowo2019adaptive}, to building a meta-classifier for sensor selection \citep{zappi2008activity}. These methods all use immediate rewards to perform feature selection. For predicting long activity sequences, this potentially ignores the information that a feature may have on future predictions, or conversely, overestimate the importance of a feature given previous observations.

Among the RL based approaches, some methods attempt to build an MDP to decide which feature to select next or whether to stop acquiring features and make a prediction \citep{he2012cost, karayev2013dynamic, kolamunna2016afv}. These methods condition the choice of one feature on the observation generated by another one, instead of choosing between all sensors simultaneously. \citet{spaan2009decision} and \citet{satsangi2015exploiting} formulated a Partially Observable MDP (POMDP) using a discretization of the continuous state to model the policy. \citet{yanginstance} formulate an RL objective by penalizing the prediction performance by the number of sensors used. Although using a desirable objective, the method employs a greedy maximization process to approximately solve the combinatorial optimization. Moreover, they do not integrate easily with existing deep architectures.

Attention is another method worth noting, as it is able to select the most relevant segments of a sequence for the current prediction \citep{vaswani2017attention}. Attention modules have been recently used for activity recognition \citep{ma2019attnsense}. However, like most attention methods, it requires all of the features to be observed before deciding which features are the most important for prediction. Moreover, the number of instances attended to is not penalized. Finally, soft attention methods typically weight the inputs, instead of selecting the feature subset. Indeed, our experiments on naively applying attention for dynamic feature selection show that it always selects 100\% of the features at all times.

Selection or skipping along the temporal direction to decide when to memorize or update the model state has been considered by \citet{hu2019learning,campos2018skip,neil2016phased}. They either are not context dependent or do not consider energy efficiency or interpretability. Additionally, skipping time steps may not be suitable for continuous monitoring tasks including HAR, where we are tasked to predict at every time step. Our dynamic feature selection is orthogonal to temporal selection/skipping and we leave exploring the potential integration of these two directions as our future research.

\subsection{Sparse Regularization}

Sparse regularization has previously been formulated for deep models, e.g., works by \citet{liu2015sparse,louizos2017learning,frankle2018lottery}. In particular, $\ell_1$ regularization is a common method to promote feature sparsity \citep{tibshirani1996regression, friedman2010note, friedman2008sparse, zou2005regularization}. However, their focus has primarily been in statically compressing model sizes or reducing overfitting, instead of dynamically selecting features for prediction.

\subsection{Variational Inference}


There have been significant efforts in variational Bayes methods, aiming at addressing the limitations of the classical mean-field variational inference (VI)~\citep{giordano2015linear}. These methods improve the mean-field posterior approximation using linear response estimates \citep{giordano2015linear, giordano2018covariances}, or adding dependencies among the latent variables using a structured variational family~\citep{saul1996exploiting}, typically tailored to particular models~\citep{ghahramani1997factorial, titsias2011spike}. Other ways to add dependencies among the latent variables are mixtures~\citep{bishop1997approximating, gershman2012nonparametric, salimans2013fixed, guo2016boosting, miller2017variational}, copulas~\citep{tran2015copula, han2016variational}, hierarchical models~\citep{ranganath2016hierarchical, tran2015variational, maaloe2016auxiliary}, or recent flow-based methods with invertible transformations of random variables~\citep{rezende2015variational, kingma2016improved, papamakarios2017masked, tomczak2016improving, tomczak2017improving, dinh2016density}. There are also spectral methods~\citep{shi2017kernel} or sampling-based methods that define the variational distribution using corresponding sampling mechanisms~\citep{salimans2013fixed}. Recently, variational inference with implicit distributions construct a flexible variational family using non-invertible mappings parameterized by deep neural networks~\citep{mohamed2016learning, nowozin2016f}. The main issue of implicit distribution variational inference is density ratio estimation, which is particularly difficult in high-dimensional settings~\citep{goodfellow2016deep,sugiyama2012density}. There are also natural-gradient methods for variational inference, with \citet{lin2019fast} extending their application to estimate structured approximations.

Semi-implicit variational inference (SIVI) combines a simple reparameterizable distribution with an implicit one to obtain a flexible variational family, and maximizes a lower bound of the Evidence Lower BOund~(ELBO) to find the variational parameters~\citep{yin2018semi}. \citet{molchanov2019doubly} have recently extended SIVI in the context of deep generative models. They use a semi-implicit construction of both the variational distribution and the deep generative model that defines the prior. This results in a doubly semi-implicit architecture that allows building a sandwich estimator of the ELBO. \citet{moens2021efficient} have more recently proposed an efficient solver for SIVI for complex datasets and posteriors. Unbiased Implicit Variational Inference (UIVI) also defines the variational distribution implicitly, directly optimizing the evidence lower bound (ELBO) rather than an approximation to the ELBO~\citep{titsias2019unbiased}.

\subsection{Discrete Variable Backpropagation}

There have been many formulations that propose to solve the issue of backpropagation through discrete random variables \citep{jang2016categorical, maddison2016concrete, tucker2017rebar, grathwohl2017backpropagation, yin2018arm}. REBAR \citep{tucker2017rebar} and RELAX \citep{grathwohl2017backpropagation} employ REINFORCE and introduce relaxation-based baselines to reduce sample variance of the estimator. However, these baselines increase the computation and cause potential conflict between minimizing the sample variance of the gradient estimate and maximizing the expectation objective. Augment-REINFORCE-Merge is a self-control gradient estimator that does not need additional baselines~\citep{yin2018arm}. It provides unbiased gradient estimates that exhibit low variance~\citep{pmlr-v108-boluki20a}, but its direct application to autoregressive or sequential setups is not addressed by \citet{yin2018arm} and leads to approximate gradients. Moreover, an exact sequential formulation will require prohibitive computation, squared in sequence length forward passes. 

\section{Additional Details and Discussions}

\begin{figure*}[t]
\centering
\subfigure[]{
\includegraphics[width=0.38\textwidth]{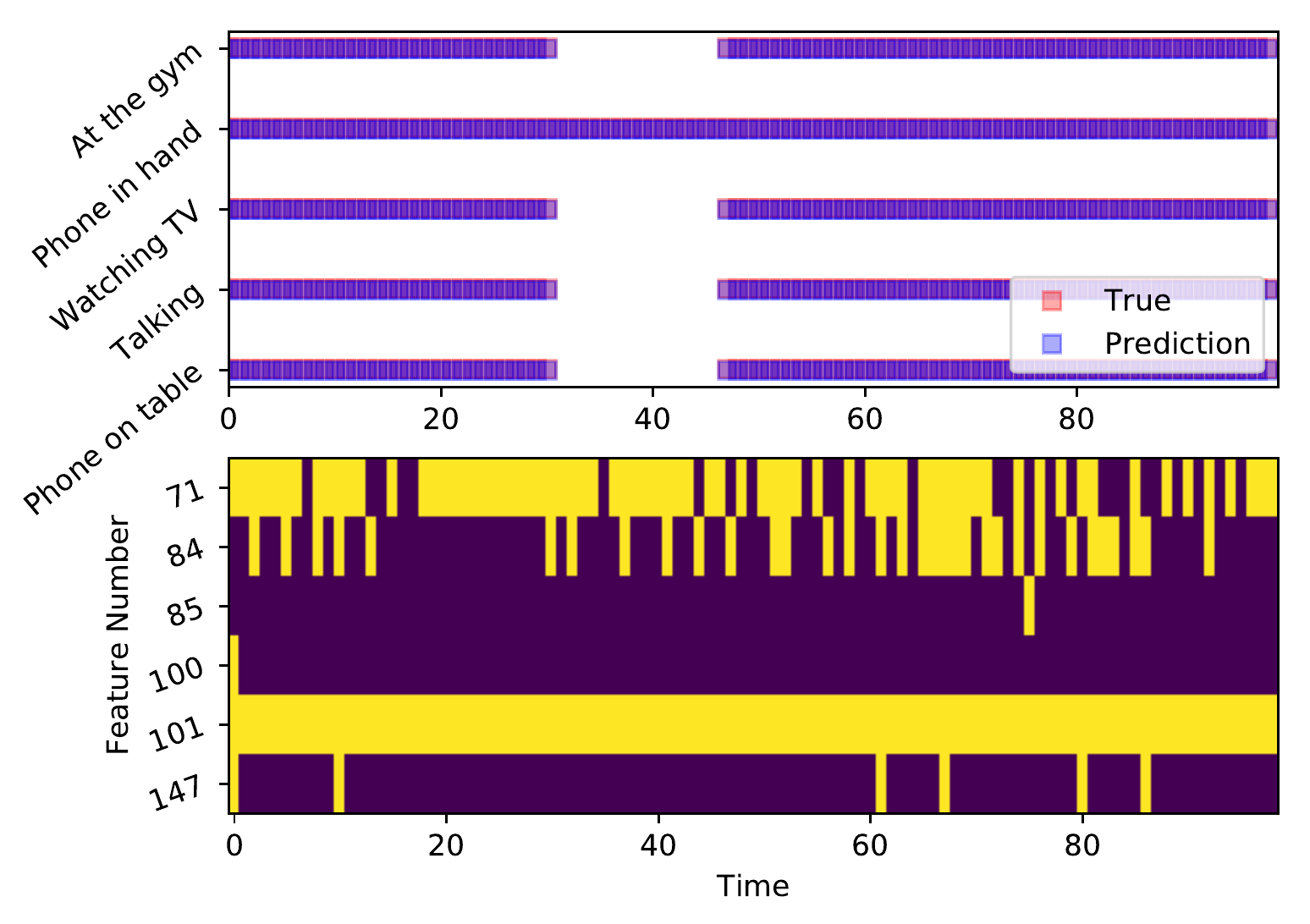}
\label{fig:traj-extra}
}
\subfigure[]{
\includegraphics[width=0.27\textwidth]{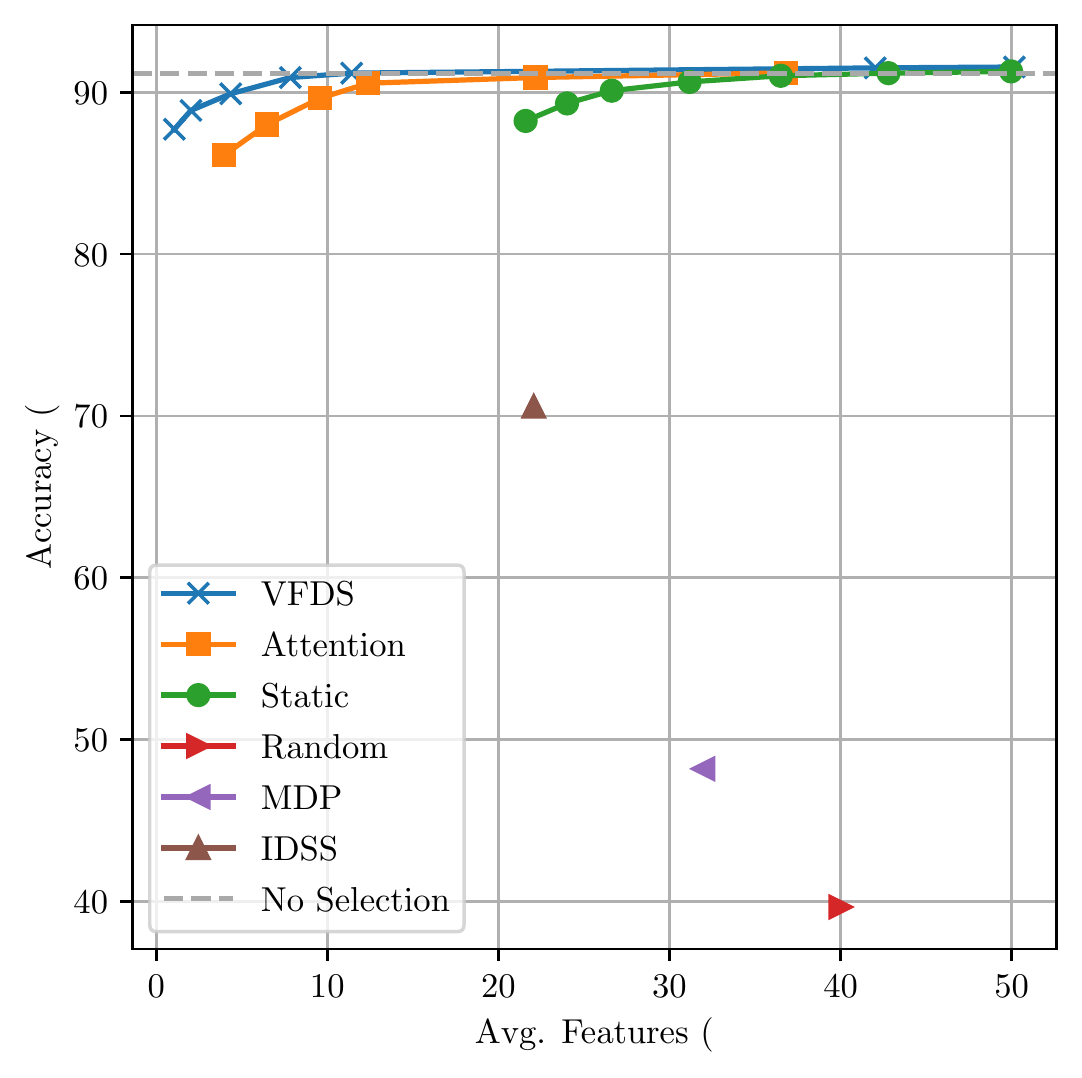}
\label{fig:tradeoff-extra-all}
}
\subfigure[]{
\includegraphics[width=0.27\textwidth]{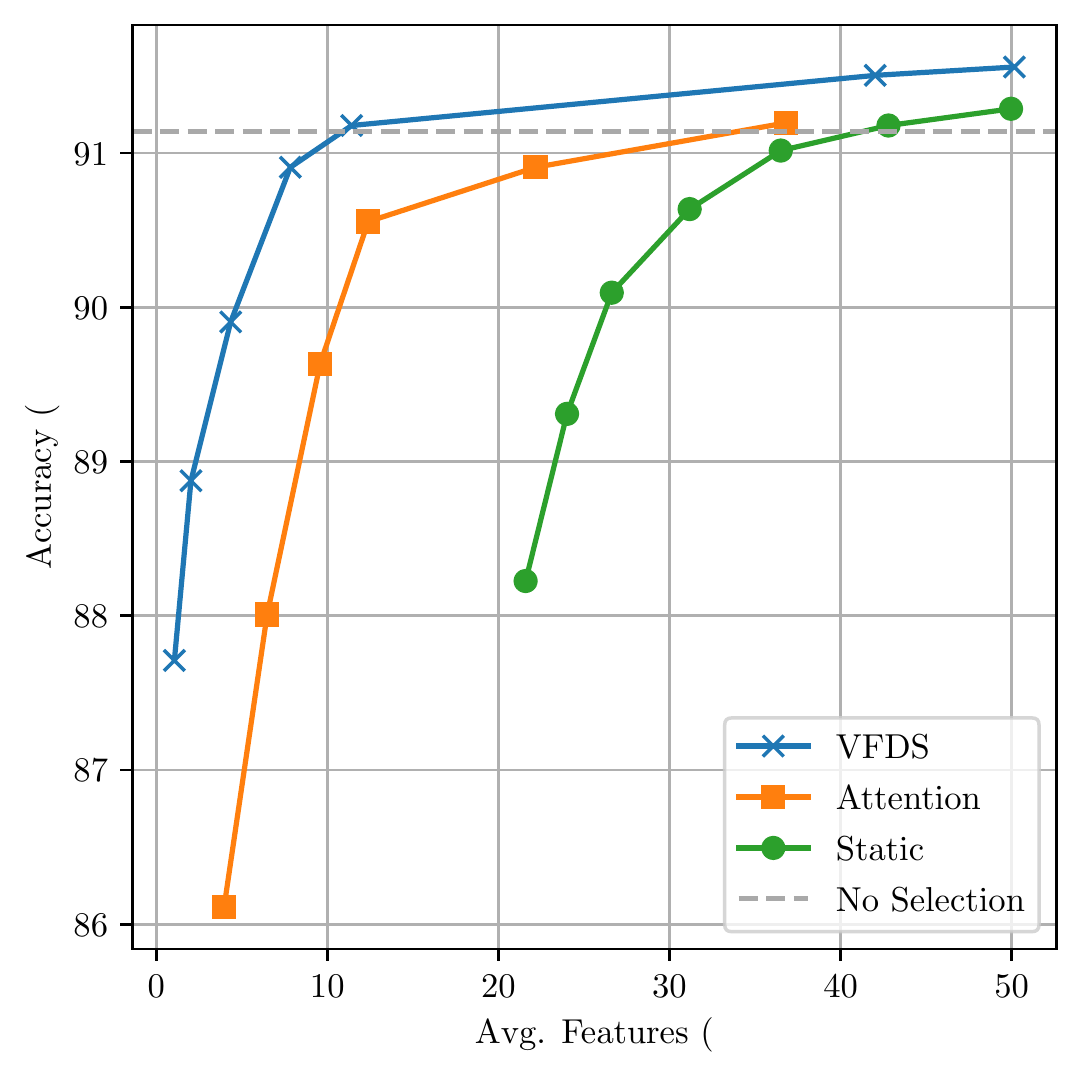}
\label{fig:tradeoff-extra}
}
\caption{ExtraSensory Dataset results: \textbf{(a)} Prediction and features selected of the proposed model. \textbf{(b)} Feature selection vs. Error trade-off curve comparison. \textbf{(c)} Feature selection vs. Error trade-off curve comparison, zoomed in on the best performing models.}
\end{figure*}

\subsection{Computational Complexity}

In general, the added computation and memory incurred by our dynamic framework consists of an additional fully connected layer used to infer the next feature set. This would only add extra $H \times P$ parameters and multiply-add operations, where $H$ is the number of hidden neurons and $P$ is the number of input features. This additional computational burden is insignificant compared to the memory and computational cost of the main network, which are typically of order higher than $O(HP)$.

\subsection{UCI HAR Dataset}

The UCI HAR dataset consists of a training set and a testing set. To implement our dynamic feature selection and other baseline methods, we divide the training set into a separate validation set consisting of 2 subjects. We preprocess the data by normalizing it with the mean and standard deviation. We then divide the instances of each subject into segments of length 200. 

The base model we utilize is a one-layer GRU with 2800 neurons for the hidden state. We use the cross-entropy of the predicted vs. actual labels as the performance measure. We use a temperature of 0.05 for the Gumbel-Softmax relaxation. We optimize this with a batch size of 10 using the RMSProp optimizer, setting the learning rate to $10^{-4}$ and the smoothing constant to 0.99 for 3000 epochs. We then save both the latest model and the best model validated on the validation set.

\subsection{OPPORTUNITY Dataset}

The OPPORTUNITY dataset consists of multiple demonstrations of different activity types. We first extract the instances into segments containing no missing labels for the mid-level gestures. Segments of length smaller than 100 are padded using the observed values at the next time-points in the instance. We then normalize the data such that its values are between -1 and 1. The authors of the dataset recommended removing some features that they believed are not useful, however we find that this does not affect performance and instead use the entire feature set. We have also experimented with interpolating the missing values but also find that it does not affect performance compared to imputing the missing values with zeros. Using this, we randomly shuffle the segments and assign 80\% for training, 10\% for validation, and 10\% for testing.

The base model we utilize is a two-layer GRU with 256 neurons for each layer's hidden state. The cross-entropy of the predicted vs. actual labels is adopted as the performance measure. We use a temperature of 0.05 for the Gumbel-Softmax relaxation. We do not include the cross-entropy loss for the time points with missing labels. We also scale the total performance loss of the observed labels for each batch by $\frac{\# \text{timepoints}}{\# \text{labelled timepoints}}$. We optimize this loss with a batch size of 100 using the RMSProp optimizer, setting the learning rate to $10^{-4}$ and the smoothing constant to 0.99 for 3000 epochs. We then save both the latest model and the best model validated on the validation set.

\subsection{ExtraSensory Dataset} 

\begin{figure*}[t]
\centering
\includegraphics[width=0.90\textwidth]{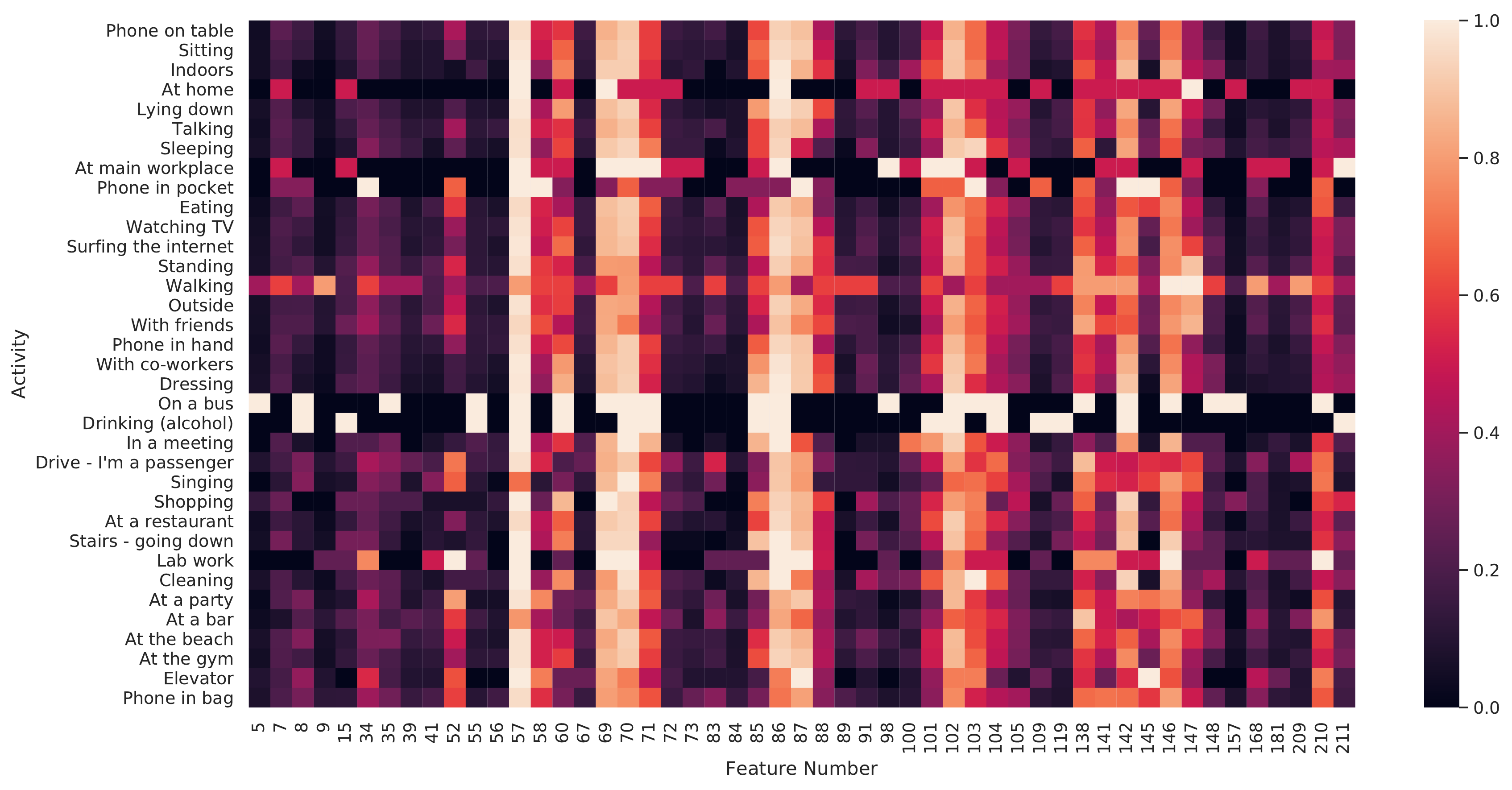}
\caption{Heatmap of sensor feature activations under each activity state of the ExtraSensory dataset.}
\label{fig:heatmap-extrasensory}
\end{figure*}

We further test our proposed method on the ExtraSensory Dataset \citep{vaizman2017recognizing}. This is a multilabel classification dataset, where two or more labels can be active at any given time. It consists of 51 different context labels, and 225 sensor features.

The ExtraSensory dataset consists of multiple demonstrations of human behavior under different activities, where two or more activity labels can be active at the same time. We first extract the instances into segments containing no missing labels for the middle level gestures. Segments of length smaller than 70 are padded using the observed values at the next time-points in the instance. We then normalize the data such that its values are in between -1 and 1. We have experimented with interpolating the missing values but also find that it does not affect performance compared to imputing the missing values with zeros. Using this, we randomly shuffle the segments and assign 70\% for training, 10\% for validation, and 20\% for testing.

We frame the problem as a multilabel binary classification problem, where we have a binary output for each label indicating whether it is active. The base model we utilize is a one-layer GRU with 2240 neurons for its hidden state. We use a temperature of 0.05 for the Gumbel-Softmax relaxation. We use the binary cross-entropy of the predicted vs. actual labels as the performance measure, where the model outputs a binary decision for each label, representing whether each label is active or not. We do not include the performance loss for the missing labels and scale the total performance loss of the observed labels for each batch by $\frac{\# \text{timepoints} \times \# \text{total labels}}{\# \text{observed labels in labelled timepoints}}$. We optimize this scaled loss with a batch size of 100 using the RMSProp optimizer, setting the learning rate to $10^{-4}$ and the smoothing constant to 0.99 for 10000 epochs. We then save both the latest model and the best model validated on the validation set.

Our method is again competitive with the standard GRU model using less than 12\% of all the features. A trade-off curve is shown in Figure~\ref{fig:tradeoff-extra-all} and Figure~\ref{fig:tradeoff-extra}, where we see a similar trend for both dynamic and attention models. However we were unable to obtain a feature selection percentage lower than 25\% for the static selection model even with $\lambda$ as large as $10^4$. We believe that this is because at least 25\% of statically selected features are needed; otherwise the static selection model will degrade in performance catastrophically, similar to the OPPORTUNITY dataset results.

A heatmap of the features selected under each activity state can be seen in Figure~\ref{fig:heatmap-extrasensory}. As shown, there are four groups of sensor features that are used across activities: the phone magnetometer (57-71), watch accelerometer magnitude (85-88), watch accelerometer direction (101-105), and location (138-147). For two particular states, `on a bus' and `drinking alcohol', phone accelerometer measurements (5-52) become necessary for prediction. Some states such as `at home', `at main workplace', and `phone in pocket' are notably sparse in sensor feature usage. We believe that these states are static, and do not require much sensor usage to monitor effectively. Other sensors such as the phone gyroscope, phone state, audio measurements and properties, compass, and various low-frequency sensors are largely unnecessary for prediction in this dataset.



\subsection{NTU-RGB-D Dataset}

\begin{figure}[h]
\centering
\includegraphics[width=0.35\textwidth]{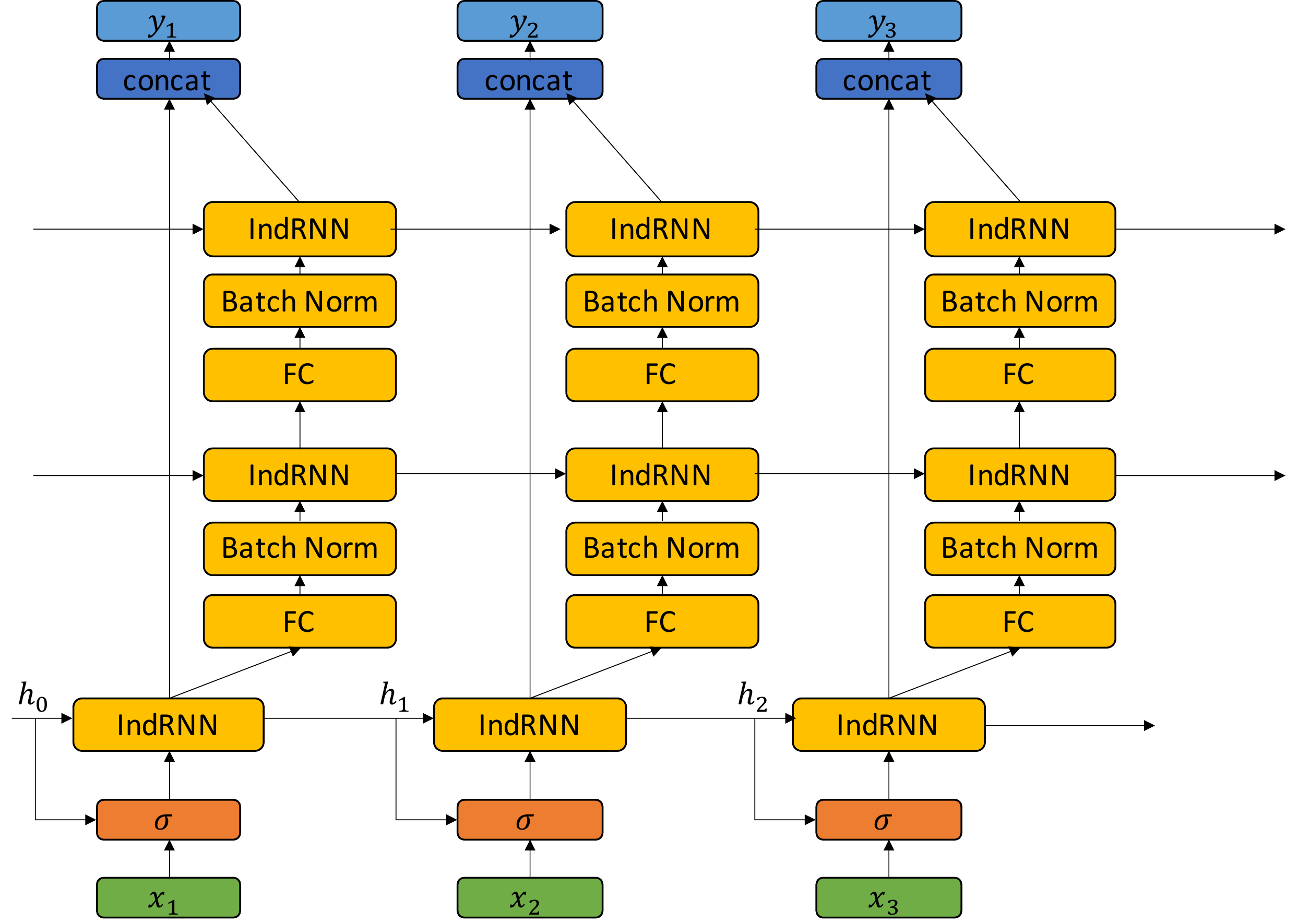}
\caption{Our modified densely connected independent RNN architecture for dynamic feature selection.}
\label{fig:architecture-ntu}
\end{figure}

We first preprocess the NTU-RGB-D dataset to remove all the samples with missing skeleton data. We then segment the time-series skeleton data across subjects into 66.5\% training, 3.5\% validation, and 30\% testing sets. The baseline model that we have implemented for the NTU-RGB-D dataset is the Independent RNN~\citep{li2018independently}. This model consists of stacked RNN modules with several additional dropout, batch normalization, and fully connected layers in between. Our architecture closely follows the densely connected independent RNN of~\citet{li2018independently}. To incorporate feature selection using either our dynamic formulation or an attention-based formulation, we add an additional RNN to the beginning of this model. This RNN takes as input the 25 different joint features and is tasked to select the joints to use for prediction further along the architecture pipeline. Since the joints are in the form of 3D coordinates, our feature selection method is modified such that it selects either all 3 of the X, Y, and Z coordinates of a particular joint, or none at all. Our architecture can be seen in Figure~\ref{fig:architecture-ntu}.

Similar as the baseline method presented by~\citet{li2018independently}, we have trained this architecture using a batch size of 128 and a sequence length of 20 using the Adam optimizer with a patience threshold of 100 iterations. We then save both the latest model and the best model validated on the validation set.

\begin{figure*}[t]
\centering
\includegraphics[width=0.70\textwidth]{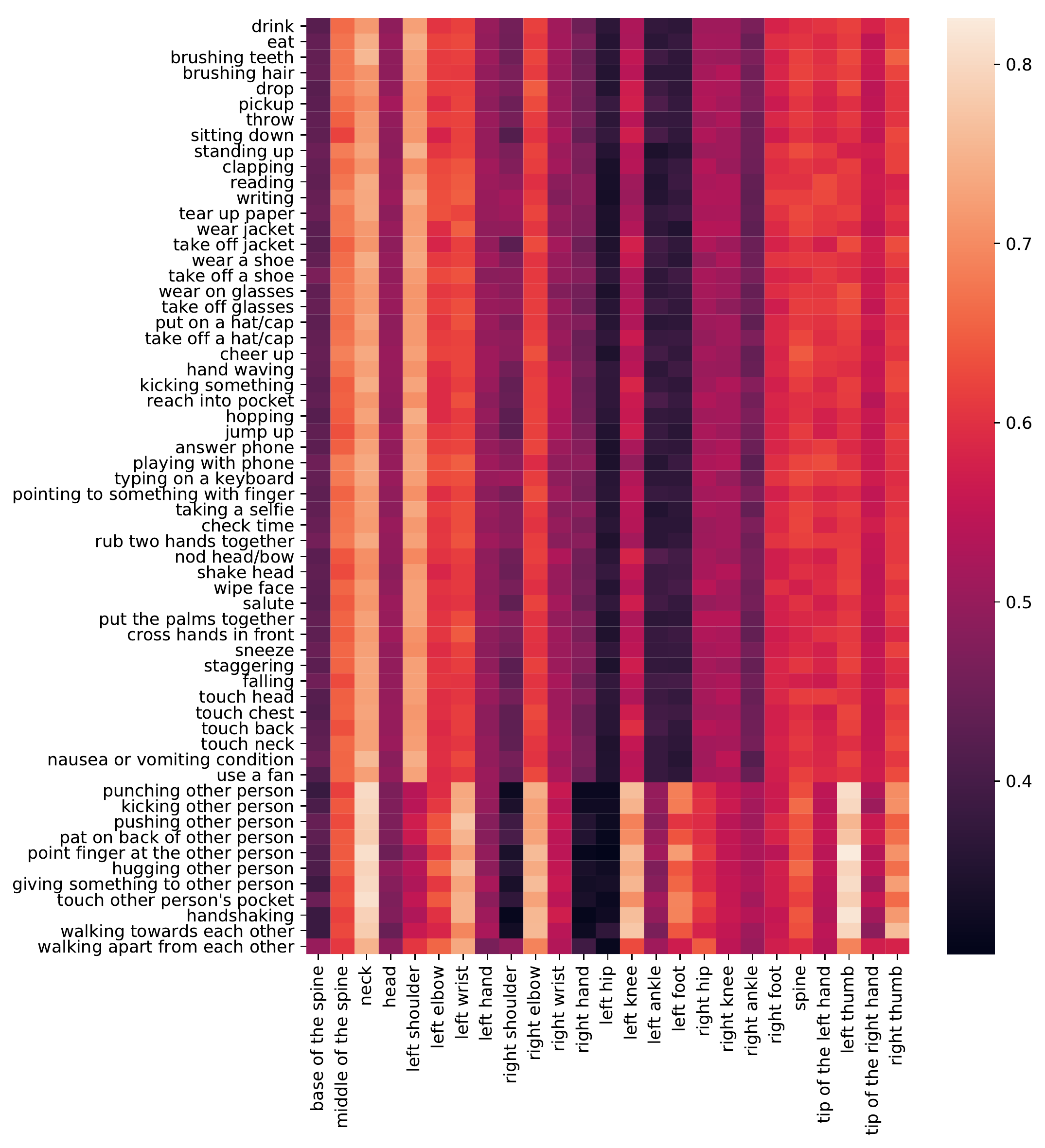}
\caption{Heatmap of sensor feature activations under each activity state of the NTU-RGB-D dataset.}
\label{fig:heatmap-ntu}
\end{figure*}

A heatmap for the features selected under each activity is shown in Figure~\ref{fig:heatmap-ntu}. Here, we can see that there are two distinct feature sets used for two different types of interactions: single person interactions and two person interactions. Indeed, since the two person activities require sensor measurements from two individuals, the dynamic feature selection would need to prioritize different features to observe their activities as opposed to single person activities. 

\section{Effect of the Temperature Hyperparameter}

\begin{figure}[h]
\centering
\includegraphics[width=0.40\textwidth]{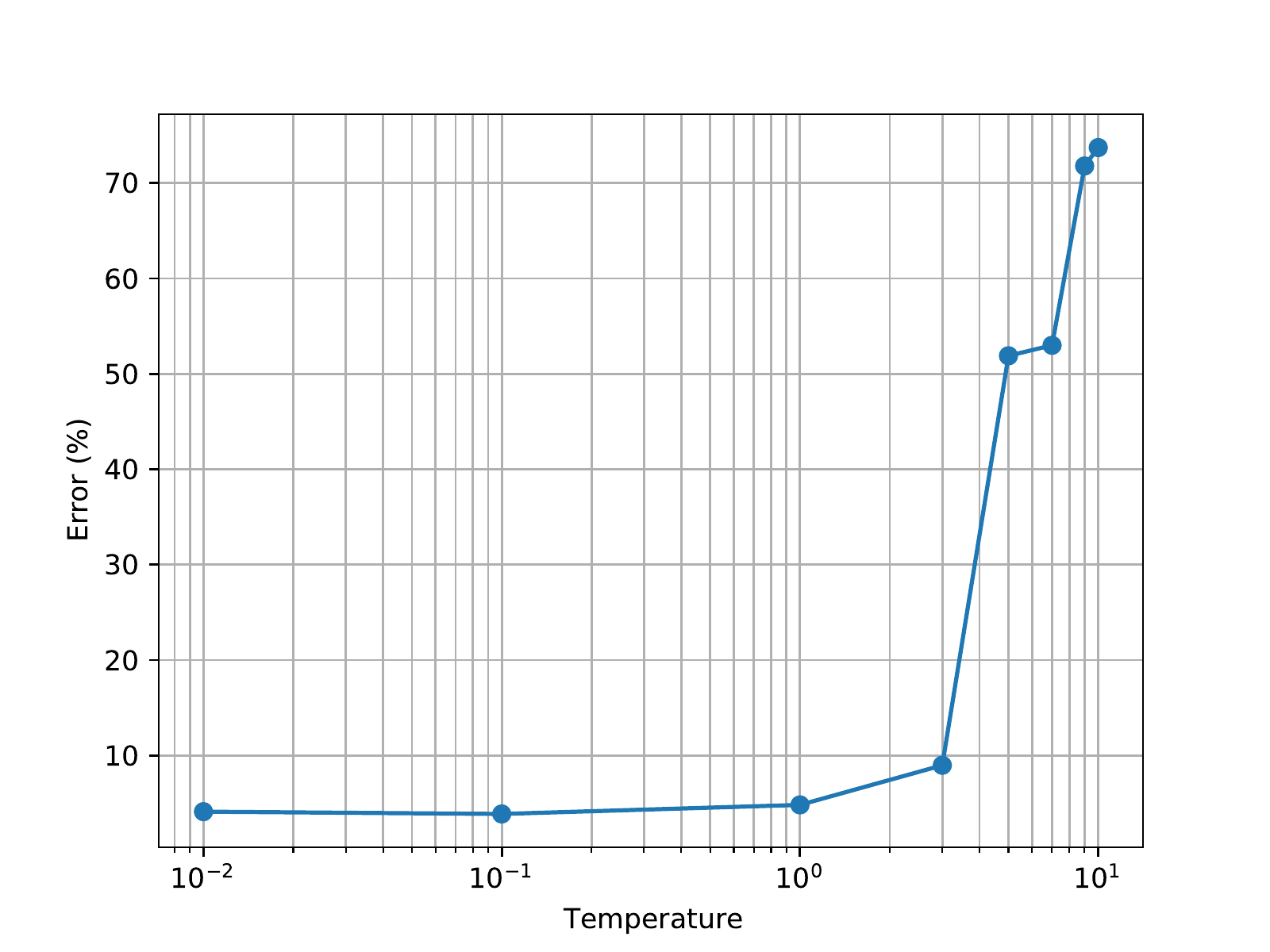}
\caption{The effect of the temperature hyperparameter $\tau$ on the performance of the model.}
\label{fig:tau-error}
\end{figure}

We further observe the effects of the temperature hyperparameter of the differentiable relaxation that we adopt on our model's performance. To do this, we have tested several hyperparameter values in our experiment with the UCI HAR dataset. The results of our tests can be seen in Figure~\ref{fig:tau-error}. In general, the settings with the temperature parameters below 1 generally yield the best results with no noticeable performance difference. Once the temperature is set to above 1, we observe a sharp increase in errors. We attribute this to the mismatch between training and testing setups, where in testing, discrete binary values are sampled while in training, the samples are reduced to an equal weighting between the features.

\section{Union of All Features Selected by the Dynamic Model}

Here, in addition to showing the average number of selected features, we compute the percentage of all features considered by our model across the full time-length. In other words, the results presented here show the union of selected features across the time horizon. In Section 4, we chose to present the average number of selected features as it directly reflects the number of required sensors for accurate HAR. Hence, it clearly shows the benefits of our proposed dynamic feature selection with respect to the power usage for sensor data collection. From Table~\ref{table:union-features}, it is clear that the percentage of all the features considered across the full time-length is also significantly low for each of the three benchmark datasets, which further validates the potential of our dynamic feature selection even when additional operational cost of turning on/off sensors needs to be considered.

\begin{table}[h]
\caption{The percentage of the union of selected features across three benchmark datasets.}
\label{table:union-features}
\begin{center}
\begin{tabular}{lc}
\toprule
Dataset & (\%) Union \\ 
\hline
UCI HAR & 3.56  \\
OPPORTUNITY & 19.83  \\
ExtraSensory & 26.66  \\
\bottomrule
\end{tabular}

\end{center}
\end{table}

\section{Model Performance and Stability Across Time}
\label{sec:moving-avg}

We show the average accuracy over every 1000 seconds of running the model on the testing subjects in the UCI HAR dataset in Table~\ref{table:moving-avg-error}. Based on the performance of the model across time, the model is shown to be stable for long-term predictions. In general, there is no clear temporal degradation in the testing performance for this dataset. Instead, the change of prediction errors is mostly dependent on the underlying activity types.



\begin{table}[h]
\caption{The average model performance across time averaged across time-aligned testing subjects.}
\vspace{-2mm}
\label{table:moving-avg-error}
\begin{center}
\begin{tabular}{l|cccc}
\toprule
Time & 0-999 & 1000-1999 & 2000-2999 & 3000-3999 \\ 
\hline
Error (\%) & 3.49 & 2.93 & 6.46 & 4.06 \\
Std. Dev. & 1.89 & 1.23 & 1.05 & 1.67 \\
\bottomrule
\end{tabular}
\end{center}
\end{table}

\bibliography{references}
\bibliographystyle{aaai}

\vfill